\documentclass{article}

\PassOptionsToPackage{numbers, compress}{natbib}

\usepackage[preprint]{tex_style}

\usepackage[utf8]{inputenc} 
\usepackage[T1]{fontenc}    
\usepackage{hyperref}       
\usepackage{url}            
\usepackage{booktabs}       
\usepackage{amsfonts}       
\usepackage{nicefrac}       
\usepackage{microtype}      
\usepackage{xcolor}         

\usepackage{graphicx}
\usepackage{subcaption}
\usepackage{booktabs} 
\usepackage{siunitx}
\usepackage{multirow}
\usepackage[version=4]{mhchem}

\usepackage{hyperref}

\usepackage{bm}
\newcommand {\pt} [1] {\left( #1 \right)}
\newcommand {\bk} [1] {\left[ #1 \right]}
\newcommand {\cb} [1] {\left\{ #1 \right\}}

\usepackage{amsmath}
\usepackage{amssymb}
\usepackage{mathtools}
\usepackage{amsthm}
\usepackage[version=4]{mhchem}

\title{Framework-Conditional Open Materials Generation}

\author{
  Maya M. Martirossyan\thanks{These authors contributed equally.} \\
  New York University\\
  \texttt{mm9084@nyu.edu} \\
  \And
  Hillary Pan$^{*}$ \\
  NYU Langone Health \\
  \texttt{hillary.pan@nyulangone.org}
  \AND
  Philipp Höllmer \\
  New York University \\
  \texttt{ph2484@nyu.edu} \\
  \And
  Stefano Martiniani\thanks{Corresponding author.} \\
  New York University \\
  \texttt{sm7683@nyu.edu}
}

\begin{document}

\maketitle

\begin{abstract}
  Theoretical predictions of novel materials rely on a variety of techniques, from data-mining-based ion substitution methods to fully \textit{de novo} generative models. Each approach imposes specific constraints and encounters unique challenges when extrapolating beyond the distribution of known materials to discover new ones. 
  The foremost goal of materials generation is to produce stable, unique, and novel (S.U.N.) structures. Predicted structures that meet these criteria, however, typically result from novel chemical compositions mapped onto known structural prototypes or frameworks.
  We leverage this heuristic insight to present a new generative paradigm for the \textit{de novo} prediction of inorganic crystalline materials constrained by their \emph{structural framework}---\textit{i.e.}, the unit cell shapes and respective particle positions.
  Our method, termed OMatG-FC (Open Materials Framework-Conditional Generation), learns \textit{via} a combination of discrete flow matching on the atomic species and stochastic interpolants for the volume scaling of the unit cell lattice, while fixing in-place the fractional coordinates and the unit-volume lattice vectors, dramatically reducing the dimensionality of the problem compared to unconstrained (framework-free) generation.
  Our method is more flexible than existing prototype-constrained methods, demonstrating capability for discovery of unseen, stable prototypes derived from known structural frameworks.
  We also investigate the number of unique frameworks in the $MP$-20 dataset, demonstrating the propensity of duplicate frameworks within the dataset and providing a framework-split version.
  We benchmark the performance of our model using the LeMat-GenBench suite using three ML interatomic potentials and achieve state-of-the-art S.U.N. rate below the convex hull, eschewing the need to learn atomic packings for materials discovery.
\end{abstract}

\section{Introduction}

A foundational goal of materials science is the design and discovery of novel materials with desired function.
The advent of battery materials \cite{kamaya_lithium_2011, jang_structurally_2024}, semiconductors \cite{nomura_room-temperature_2004}, and alloys \cite{gludovatz_exceptional_2016} have historically shaped the fabric of our society, with further technological leaps promised by ``holy-grail'' materials such as room-temperature superconductors \cite{boeri_2021_2022, prakash_guided_2025}.

Theoretical predictions of novel materials can support the discovery process: Traditional approaches include high-throughput \cite{jain_highthroughput_2011, curtarolo_highthroughput_2013} or ab-initio random structure search \cite{pickard_initio_2011}, which are based on first-principles methods like density functional theory (DFT), but suffer from poor sampling and high computational cost.
Given the vast search space of candidate materials, machine-learning (ML) methods offer a promising avenue for dramatically accelerating materials discovery. Among these, generative models \cite{de_breuck_generative_2025, metni_generative_2025, bone_discovery_2025} stand out for their ability to propose novel materials by learning from experimentally or computationally derived crystal structures represented as unit cells.

Structure is of chief importance for material properties, especially for inorganic crystals, whose behavior depends sensitively on atomic-level periodicity \cite{vu_towards_2023}. This has motivated extensive work on crystal structure prediction (CSP) from chemical composition, as well as \textit{de novo} generation (DNG) of novel compositions and their structural packings.
The primary challenge for such methods is learning to represent and sample the space of stable crystalline materials---spanning all possible structural packings and chemical species---in order to discover viable candidate materials that can be synthesized \cite{cheetham_artificial_2024}.

\subsection{Related Works}

Significant prior work in materials prediction leverages knowledge of existing materials combined with heuristic constraints.
Such methods predate application of machine-learning techniques: For example, the empirical Hume-Rothery rules \cite{hume-rothery_structure_1939} identify plausible solid solutions via substitutions of chemical species within an existing crystal structure template. 
Probabilistic models have also been applied to predict new compounds from structural templates \textit{via} ionic substitutions \cite{hautier_data_2011a}: A key feature of this method is the constraint for candidate structures to remain within their \emph{structural prototype} (see Sec.~\ref{sec:definitions}).

In contrast, machine-learned CSP models predict stable atomic packings given a fixed chemical composition. Several such CSP methods have been proposed with the goal of improving on \textit{ab initio} random structure search \cite{pickard_airss_2016}, such as CDVAE \cite{xie_crystal_2022}, DiffCSP \cite{jiao_crystal_2023}, FlowMM \cite{miller_flowmm_2024}, and OMatG \cite{hoellmer_open_2025a}. 
These methods use equivariant architectures and a range of generative frameworks---from variational autoencoders to stochastic interpolants \cite{albergo_stochastic_2023} (encompassing diffusion models \cite{song_scorebased_2021} and flow-matching \cite{albergo_building_2023, lipman_flow_2023}).
A separate line of work addresses unconstrained \textit{de novo} generation, predicting both composition and structure jointly; this includes GNoME \cite{merchant_scaling_2023} and MatterGen \cite{zeni_mattergen_2023}, as well as variants of several methods mentioned above \cite{hoellmer_open_2025a, miller_flowmm_2024}.
Chemeleon2 \cite{Park2025chemeleon2} extends this by fine-tuning a latent diffusion model with group relative policy optimization.
Diffusion transformer have been explored as a more lightweight architectural alternative: ADiT \cite{joshi_allatom_2025} uses a shared latent space and diffusion transformer for jointly generating materials and molecules, while Crystalite \cite{veljkovic_crystalite_2026} uses a diffusion transformer whose attention is augmented with periodic pairwise geometric information.

Another class of ML methods imposes physically-motivated inductive biases by exploiting the redundancy inherent in crystalline symmetries. 
Crystals possess translational and point-group symmetries that together describe a space group, which generates sets of symmetry-equivalent positions within the  unit cell. 
When occupied by atoms, these positions are known as Wyckoff positions. Each Wyckoff position specifies both the multiplicity (how many symmetry-equivalent atoms it generates) and the degrees of freedom for atomic coordinates. 
For a given space group, operating directly on Wyckoff positions rather than on all atomic coordinates simplifies the learning objective by removing redundant degrees of freedom.

DiffCSP++ \cite{jiao_space_2024} takes this approach, requiring Wyckoff templates as input and learning only the continuous coordinates within those sites. Since ground-truth symmetry is unavailable, templates must be retrieved from the training set based on compositional similarity, restricting generation to known structural prototypes. SymmCD \cite{levy_symmcd_2024}, CrystalFormer \cite{cao_space_2025}, and SGEquiDiff \cite{chang_space_2025} instead incorporate Wyckoff site selection as a discrete generative step, enabling the discovery of structural prototypes beyond those seen during training.
PLaID++ \cite{xu_plaid_2025} also uses a Wyckoff-based text representation, applying direct preference optimization to steer a language model.
Broadly, such symmetry-constrained models typically show improved space group distributions of predicted structures---towards more symmetric crystal structures---which in turn improves stable, unique, and novel (S.U.N.) rates.

Finally, SCIGEN \cite{okabe_structural_2025a} takes a complementary approach, steering diffusion-based generation toward desired lattice motifs (such as kagome or honeycomb) by constraining the denoising process, and has led to the successful synthesis of two predicted materials.

\subsection{Our Contribution}\label{sec:our_contribution}

It is well-understood in materials science that two crystal structures sharing the same \textit{structural framework} can exhibit different space group symmetries. 
A structural framework specifies the unit cell shape and the fractional coordinates of atomic sites, but not the chemical species occupying each site (see Sec.~\ref{sec:definitions} and Fig.~\ref{fig:bcc_CsCl}).
In fact, partitioning sites among distinct species can reduce the symmetry to a subgroup of the framework's full geometric symmetry.\footnote{A simple demonstration of this difference in symmetry can be seen between a ring of identical beads (invariant under any rotation $2\pi/N$ where $N$ is the number of beads) and a ring of beads of alternating color (invariant under any rotation $2\pi/(N/2)$ where $N$ is the number of beads and $2$ is the number of colors). The two-color ring of beads is less symmetric than the one-color ring, which has the full geometric symmetry.} 

It is unknown whether symmetry-constrained generative models might benefit from operating at the structural framework level. 
Current models produce few novel structural prototypes \cite{szymanski_2025_matgen_baselines}, and it remains unclear whether this limited novelty reflects shortcomings of the models or instead the fact that nature has already explored most viable frameworks. 
Indeed, the relatively small number of known structural prototypes---2050 entries in the AFLOW Prototype Encyclopedia \cite{eckert_aflow_2024a} at the time of writing and $\approx$\num{15000} additional prototypes in the AFLOW database \cite{hicks_aflowxtalfinder_2021}---suggests that the space of geometrically stable arrangements may be highly constrained. 
This motivates an alternative approach: rather than learning to generate novel atomic coordinates, one can leverage known frameworks and focus on predicting which species assignments yield stable materials.

\begin{figure}[t]
   \centering
   \includegraphics[width=0.6\columnwidth]{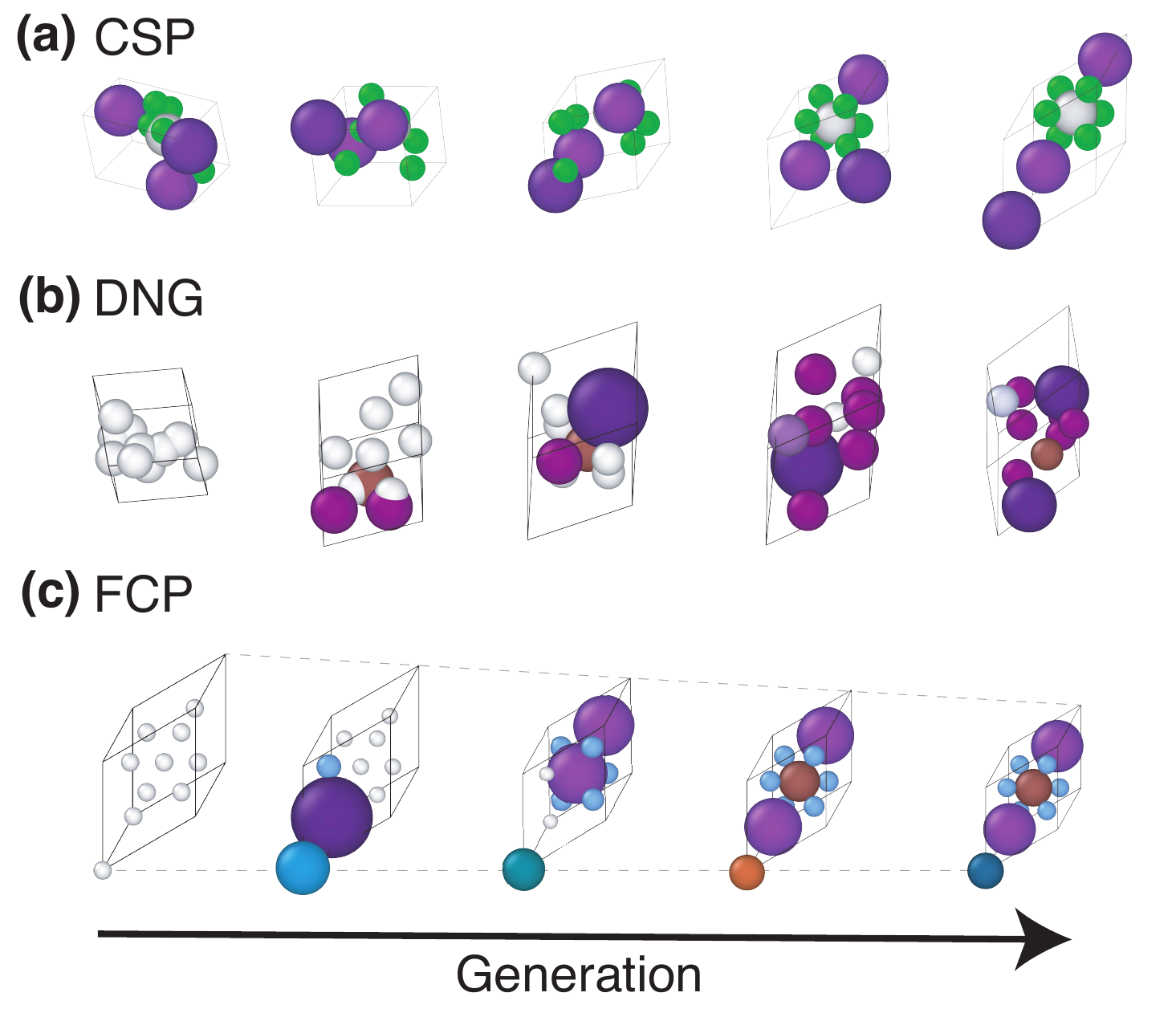}
   \caption{Illustration of the generation process with different types of generative models for the same test structure: \textbf{(a)} crystal structure prediction (CSP) samples the species at time $t=0$ from the structure at time $t=1$; \textbf{(b)} \textit{de novo} generation (DNG) evolves the lattice, species, and fractional coordinates jointly; \textbf{(c)} framework-conditional prediction (FCP) evolves the lattice volume scaling and atomic species only.}
   \label{fig:csp_dng_fcp}
\end{figure}

We introduce OMatG-FC, a new `framework-conditional' paradigm for generative models of inorganic crystals, which conditions generation on structural frameworks drawn from known crystalline materials.
Unlike prototype-constrained methods, OMatG-FC fixes only the framework geometry; any additional realized chemical decorations or symmetry-breaking emerge from species assignment rather than being imposed. This places our approach between symmetry-based methods, which allow geometric variation within space groups, and substitution-based methods, which constrain candidate species on a known structural prototype.
We demonstrate in Figure~\ref{fig:csp_dng_fcp} how our model, which performs framework-conditional prediction (FCP), differs from standard CSP and DNG models.

We show that our model learns the symmetries common in the dataset from species-to-site assignment by good performance across distributional benchmarks and S.U.N. metrics.
Moreover, our model demonstrates superior ability to predict stable, novel, and unique (S.U.N.) structures. 
While not the focus of this paper, our approach is well-suited for the prediction of chemical disorder on crystallographic sites, an active area of research particularly for materials such as multi-component alloys \cite{jakob_learning_2026, antypov_classification_2025}.

\section{Background}

\subsection{Crystal Structure Definitions}\label{sec:definitions}

A crystal structure is defined by its translational symmetry, meaning it can be represented compactly by a periodic repeating subunit---termed a unit cell.
Because of the constraint of translational symmetry combined with point-group symmetry, crystals in three dimensions must obey the symmetries belonging to one of 230 space groups.
An infinite number of crystal structures can exist within each space group, which defines the possible symmetry-inequivalent sites, known as Wyckoff positions, that can be occupied.
To specify a particular crystalline material, one needs to define its unit cell: the particle positions, the chemical species occupying each position, and the lattice parameters.
Defining all particle positions, however, is equivalent to defining the space group and all occupied Wyckoff positions, along with the choices of the free parameters (if extant) within them.

We define some terms that will be used throughout this manuscript.\footnote{The language we define may not be universal.} The structural \emph{prototype} is the \textit{discrete} classification of a crystal structure by its space group, occupied Wyckoff positions, and stoichiometry. A prototype is a structural class: many materials with different chemical species and lattice parameters can belong to the same prototype. The AFLOW Prototype Encyclopedia \cite{eckert_aflow_2024a, hicks_aflowxtalfinder_2021} follows this convention, encoding prototypes with labels such as \texttt{AB4C3\_oI32\_46\_b\_2bc\_bc}, which specify the stoichiometry, Pearson symbol (combining the Bravais lattice and the number of atoms per conventional unit cell), space group number, and Wyckoff positions, respectively. AFLOW further indexes representative materials within each prototype using numerical suffixes (\texttt{-001}, \texttt{-002}), with each entry specifying the lattice parameters and Wyckoff coordinates for that material. A crystalline \emph{material} is thus a specific instantiation of a prototype: it specifies which chemical species occupy each stoichiometric site (e.g., A~=~Ca, B~=~Ti, X~=~O for the $ABX_3$ perovskite CaTiO$_3$) along with the lattice parameters and atomic coordinates.

\begin{figure}[t]
   \centering
   \includegraphics[width=0.5\columnwidth]{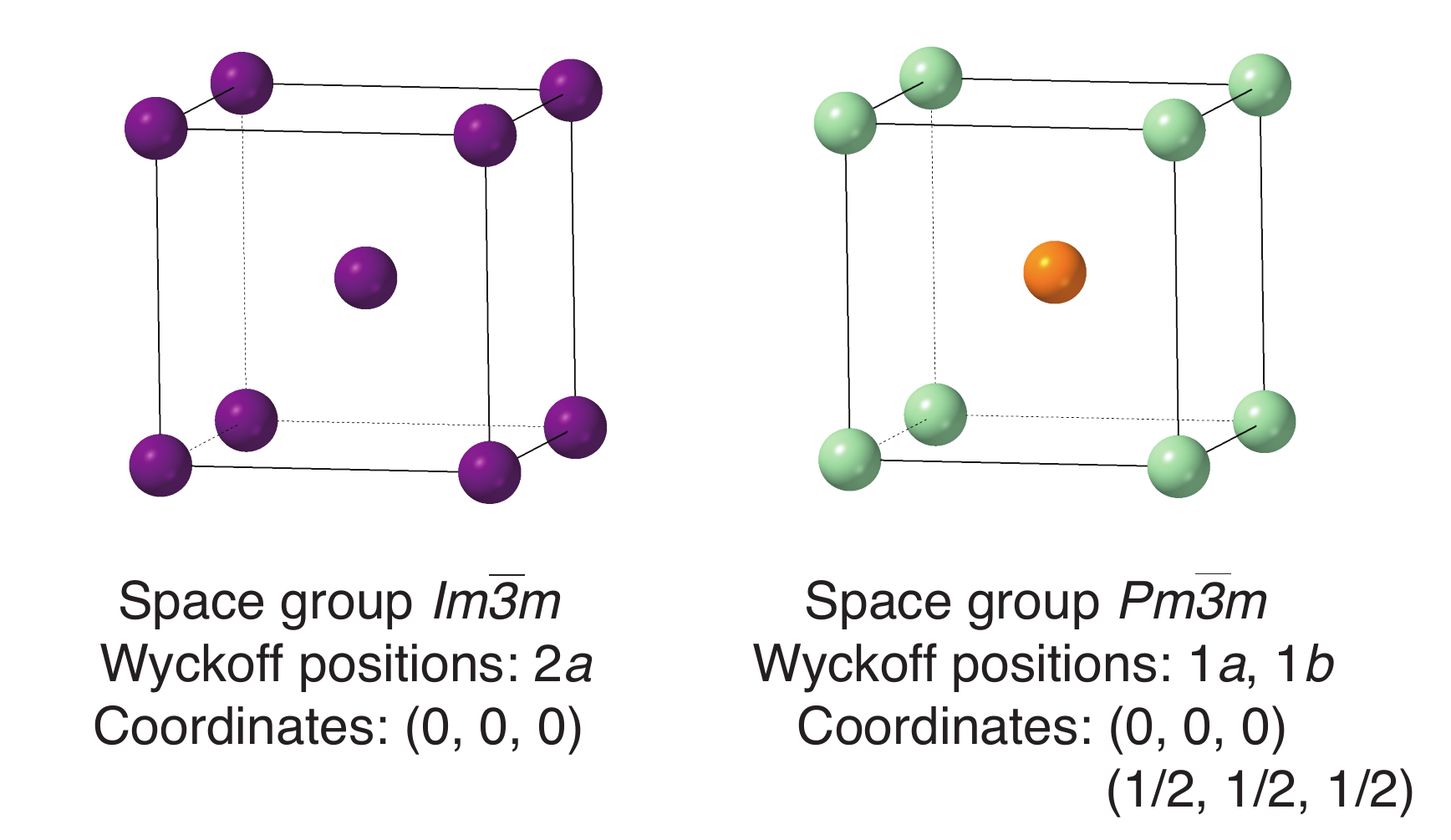}
   \caption{Two crystal structures---a BCC-type crystal (left) and a CsCl prototype (right)---with the same structural framework but different structural prototypes, labeled by their space groups and corresponding Wyckoff positions.}
   \label{fig:bcc_CsCl}
\end{figure}

We define a structural \textit{framework} as the geometric specification of a crystal independent of chemical identity: the unit cell shape (lattice parameters up to uniform scaling) together with the fractional coordinates of all atomic sites.  A framework does not encode which species occupy these sites. 
Therefore, the same framework can correspond to multiple prototypes depending on species assignment. 
For example, BCC elemental metals (space group $Im\bar{3}m$, Wyckoff site 2$a$, stoichiometry A) and CsCl (space group $Pm\bar{3}m$, Wyckoff sites 1$a$ + 1$b$, stoichiometry AB) share identical frameworks but constitute different prototypes, because the body-centered translation is a valid symmetry only when both sites are chemically equivalent (see Fig.~\ref{fig:bcc_CsCl}).

\subsection{Stochastic Interpolants}

Stochastic interpolants \cite{albergo_building_2023, albergo_stochastic_2023} unify diffusion models \cite{song_scorebased_2021} and conditional flow-matching \cite{albergo_building_2023, lipman_flow_2023} under a single framework, bridging a base probability distribution $\rho_0$ with the true data distribution $\rho_1$ through samples ($x_0 \sim \rho_0, x_1 \sim \rho_1$).
We consider interpolants of the form
\begin{equation}
x(t, x_0, x_1, z) = \alpha (t) x_0 + \beta (t) x_1 + \gamma (t) z,
\label{eq:loss_si}
\end{equation}
where $t \in [0,1]$ represents time such that $x(t=0) = x_0$ and $x(t=1) = x_1$.
A learned time-dependent drift $b^{\theta}(x,t)$ and, optionally, a learned denoiser $z^\theta(x,t)$ give access to trajectories from the time-dependent density $\rho_t$ through ODE-based or SDE-based sampling starting from $x_0\sim\rho_0$. 
For training SDE-based sampling schemes, $\gamma (t) \neq 0 $ for $t \in (0, 1)$ is required, while it is optional for ODE-based sampling.

For both ODE- and SDE-based sampling, the required velocity term $b^\theta(t, x)$ is learned by minimizing the loss function
\begin{equation}
        \mathcal{L}_b(\theta) = 
         \, \mathbb{E}_{t, z, x_0, x_1}
         \big[ |b^\theta(t, x_t)|^2 - 2\,\partial_t x(t, x_0, x_1, z) \cdot b^\theta(t, x_t) \big],
\label{eq:loss_b}
\end{equation}
with the expectation taken independently over ${t\sim \mathrm{Uniform}\bk{0,1}}$, $z\sim\mathcal{N}(0, \bm{I})$, $x_0\sim\rho_0$, and $x_1\sim\rho_1$.
For SDE-based sampling, an additional denoiser $z^\theta(t,x)$  must be learned by minimizing the additional loss
\begin{equation}
    \mathcal{L}_z(\theta) = \mathbb{E}_{t, z, x_0, x_1} \left[ |z^\theta(t, x_t)|^2 - 2\,z^\theta(t, x_t) \cdot z \right].
\label{eq:loss_z}
\end{equation}
We note that minimizing with respect to these loss functions is equivalent to minimizing a mean-squared error loss.

Prior work has explored stochastic interpolants in the context of crystal structure prediction and \textit{de novo} materials generation in the OMatG model \cite{hoellmer_open_2025a}. Following this and related work \cite{miller_flowmm_2024, bose_se3stochastic_2024}, we apply inference-time velocity annealing which rescales the learned velocity field as $b^\theta(t,x) \rightarrow (1+st)\,b^\theta(t,x)$, where $s$ is as an hyperparameter that increases the magnitude of the velocity with time.

\subsection{Discrete Flow Matching}

Discrete Flow Matching (DFM) \cite{campbell_generative_2024,gat_discrete_2024} is a generative modeling framework for categorical variables which extends flow matching to discrete state spaces. 
Similarly to stochastic interpolants, DFM relies on connecting samples $d_0$ from the (discrete) base distribution to data samples $d_1$ from the data distribution with a conditional flow. 
Types of base distributions for sampling $d_0$ include uniform and completely masked.
For the uniform conditional flow, the probability for the discrete variable $d_t$ at time $t$ given the target sample $d_1$ is given by $p_{t|1}^\text{unif}(d_t|d_1)=\mathrm{Cat}(t\,\delta\{d_1,d_t\}+(1-t)/S)$, where $S$ is the number of possible states. This uniform conditional flow converges to $d_1$ at $t=1$ and is uniformly distributed over all possible states at $t=0$. Alternatively, the masked conditional flow is given by $p_{t|1}^\text{mask}(d_t|d_1)=\mathrm{Cat}(t\,\delta\{d_1,d_t\}+(1-t)\,\delta\{M,d_t\})$, which converges to an additional masking token $M$ at $t=0$.

The conditional flows can be used to train a neural network to predict the denoising probability $p_{1|t}^\theta(d_1|d_t)$ of the final state $d_1$ given the current state $d_t$ by minimizing a cross-entropy loss:
\begin{equation}
    \mathcal{L}_{\mathrm{DFM}}(\theta) = \mathbb{E}_{t, d_{1}, d_{0}} \left[ - \log p_{1|t}^\theta(d_1|d_{t}) \right].
\label{eq:loss_dfm}
\end{equation}
Once the denoising probability $p^\theta_{1|t}(d_1|d_t)$ is learned, one can construct the rate matrix $R_t$ that determines the rate with which state $d_t$ jumps to another state at time $t$ during generation~\cite{campbell_generative_2024,hoellmer_open_2025a}. In addition, it is possible to add stochasticity in the shape of a second rate matrix satisfying detailed balance, $R_t^{DB}$, with a tunable noise parameter $\eta$ such that $R_t^\eta = R_t + \eta R_t^{DB}$.

\section{Crystal Representation and Generation}

A crystalline unit cell defines a distinct choice of periodic repeating unit which can be tiled to generate the infinite crystal structure.
Generative models typically represent a unit cell with $N$ atoms using three components: chemical species $\bm{A} \in \mathbb{Z}_{>0}^N$, lattice vectors $\bm{L} \in \mathbb{R}^{3 \times 3}$, and fractional coordinates $\bm{F} \in [0, 1)^{3\times N}$. In this work, we further decompose the lattice vectors into unit-volume lattice vectors $\bm{\bar{L}}  \in  \mathbb{R}^{3 \times 3} \, \mathrm{where} \,\det(\bm{\bar{L}}) = 1$ and a log-volume scale factor $V = \log |\det{(\bm{L})}|$.
 
Our generative modeling framework seeks to generate samples from the joint conditional distribution $\rho_1\pt{\bm{A}, V | \bm{\bar{L}}, \bm{F}}$ where $\bm{\bar{L}}$ and $\bm{F}$ define a structural framework that is sampled from the dataset.
We sample this joint distribution by co-evolving $V$ by stochastic interpolants and $\bm{A}$ by discrete flow matching, with combined loss 
\begin{equation}
    \mathcal{L}(\theta) = \lambda_{b}^V \mathcal{L}_b(\theta) + \lambda_{z}^V \mathcal{L}_z(\theta) + \lambda^{\bm{A}} \mathcal{L}_{\mathrm{DFM}}(\theta)
\label{eq:loss_tot}
\end{equation}
where $\lambda_{z}^V = 0$ for ODE sampling schemes. Relative weights $\lambda$ are optimized via hyperparameter tuning (see Appendix~\ref{app:hyperparams}).
\subsection{Chemical Species}

We represent chemical species by atomic number and model them using discrete flow matching, following OMatG's implementation.
Each atom takes a value $\bm{a}_i \in \cb{1, 2, \dots, 100}$ and $M=0$ is a mask token. The base distribution sets all atoms to the masked state $\rho_0(\bm{A}) = \bk{M}^{N}$ and atomic identities evolve \textit{via} a continuous-time Markov Chain and are progressively unmasked until all tokens are replaced at time $t=1$.
The learned denoising probability $p^\theta_{1|t}$, and hence the rate matrix, depend on all other state variables ($\bm{F}$, $\bm{\bar{L}}$, $V$) in addition to the atom types.

\subsection{Lattice Vectors and Volume Scale}

The lattice matrix $\bm{L}$ is defined by the three lattice vectors $[\bm{l}_1, \bm{l}_2, \bm{l}_3]$ and can be decomposed into shape and scale: the unit-volume lattice $\bm{\bar{L}}$ and log-volume $V$, related by $\bm{\bar{L}} = \bm{L} * \exp{\pt{-V / 3}}$.
We fix $\bm{\bar{L}}$ throughout generation using an identity interpolant with mirrored based distribution $\bm{\bar{L}}_t = \bm{\bar{L}}_0 = \bm{\bar{L}}_1$, meaning that samples $x_0$ are taken to have equivalent $\bm{\bar{L}}$ to the sampled structures $x_1 \sim\rho_1$.

For the volume scale $V$, we utilize a linear interpolant where $\alpha(t) = 1-t$ and $\beta(t) = t$ with both an ODE $\pt{\gamma(t) = 0}$ and SDE sampling scheme ($\gamma(t) = \sqrt{at\pt{1-t}}$ where $a$ is a tunable parameter).
As the stochastic interpolant framework permits flexible base distributions, we fit a log-normal distribution to unit cell volumes in the training set, yielding a Gaussian prior $\rho_0(V)$ on log-volume; previous work has shown that such priors improve generation performance~\cite{miller_flowmm_2024, hoellmer_open_2025a}.

\subsection{Atomic Coordinates}

Atomic positions live on a torus and are represented as fractional coordinates $\bm{F} \in [0,1)^{3 \times N}$, expressing each atom's position relative to the lattice vectors. As with the unit-volume lattice, we fix fractional coordinates throughout generation using an identity interpolant with mirrored base distribution, setting $\bm{F}_t = \bm{F}_1$ for all $t$, meaning that samples $x_0$ are taken to have equivalent $\bm{F}$ to the sampled structures $x_1 \sim \rho_1$.

\section{Datasets}\label{sec:datasets}

For this work, we utilize two datasets: the \textbf{\textit{MP}-20} \cite{jain_commentary_2013, xie_crystal_2022} dataset from the Materials Project that contains \num{45229} structures with at maximum $N=20$ atoms per unit cell and the \textbf{\textit{MPTS}-52} \cite{baird_matbenchgenmetrics_2024} dataset also from the Materials Project containing \num{40476} with at maximum $N=52$ atoms per unit cell and organized chronologically (time-split).
The $MP$-20 dataset utilizes a \num{60}--\num{20}--\num{20} \% train--validation--test split while $MPTS-52$ has an approximately \num{68}--\num{12}--\num{20} \% split ratio.



\subsection{Framework Duplicate Analysis}\label{sec:deduplication}


We analyze structural framework duplication within $MP$-20 to motivate our framework-conditional approach. Following \citet{martirossyan_all_2025}, we use pymatgen's \texttt{StructureMatcher} with the species-agnostic \texttt{FrameworkComparator} and volume rescaling enabled, restricting our analysis to the site tolerance \texttt{stol} (see Appendix~\ref{app:SM}). For each structure pair, we identify the smallest \texttt{stol} at which the two frameworks transition from matching to non-matching; by construction, this value equals the per-atom RMSD between aligned atomic positions returned by \texttt{StructureMatcher}. Adopting the default tolerance \texttt{stol}$=0.3$ as the matching threshold, only ${\sim}10\%$ of structural frameworks in $MP$-20 are unique. This pervasive framework redundancy across chemically distinct materials provides direct empirical support for the inductive bias underlying OMatG-FC. Further details are provided in Appendix~\ref{app:framework_splits}.


\subsection{Framework-Split Dataset}

We utilize the structural comparisons of frameworks within $MP$-20 to create a framework-split version of the dataset,  utilizing \texttt{StructureMatcher} as described in Appendix~\ref{app:deduplication} (\texttt{stol}$=0.3$) in order to identify groupings of structures by their structural similarity.
We choose this more lenient metric as to make the framework-split \emph{more} challenging: at test time, crystal structures which are the most dissimilar to the training set are held out, and performance is measured by the model's ability to perform materials discovery on previously unseen frameworks---in short, to predict novel chemistry on novel structural frameworks.
The procedure to create the split is described in greater detail in Appendix~\ref{app:fs-code}.
We provide benchmarked results on this split in Appendix~\ref{app:fs-bench}.

\section{Methodology}

\subsection{Equivariant Model for Crystal Structures}

In order to properly account for symmetries inherent to crystal structures, we utilize a CSPNet architecture based on that of \citet{jiao_crystal_2023} in the DiffCSP generative model.
CSPNet is translation-invariant as well as rotation- and permutation-equivariant, meaning it correctly accounts for the symmetries which emerge as a result of treating the unit-cell as a periodic object.
The full lattice matrix $\bm{L}$ is reconstructed and provided to the model as input.
The architecture uses learnable atomic embeddings and represents fractional coordinates through sinusoidal positional embeddings. 
It uses six message-passing layers with a fully-connected graph neural network with atomic positions as nodes.
Further details are provided in Appendix~\ref{app:CSPNet}.

\subsection{De Novo Generation}\label{sec:dng}

We benchmark OMatG-FC as a \textit{de novo} generative model in order to quantify whether conditioning on crystalline frameworks provides a strong prior for materials discovery. We evaluate the model in two modalities: first, treating framework-conditioning as a hard constraint by evaluating each generated structure as-is; second, relaxing this constraint by postprocessing with structural relaxation using the Universal Model for Atoms (UMA) uma-s-1.1 machine-learned interatomic potential (MLIP) \cite{wood_uma_2025}.
Both modalities leverage the same structural priors, but differ in whether those priors are enforced as constraints on the final output.

\subsection{Comparisons to Other Models}\label{sec:other_models}

We compare OMatG-FC's capabilities for materials discovery on the  \textit{de novo} generation task against several types of models, including site-substitution models as a reduced-task baseline, symmetry-constrained or space group conditional models, and fully \textit{de novo} generative models.
Additional details on the comparison to prototype-constrained models such as random enumeration and ionic substitution are found in Appendix~\ref{app:MGB}.
These comparisons should be interpreted primarily with respect to the (M.)S.U.N. benchmark, as our training task differs from others in the existing literature.

\section{Benchmarks}\label{sec:benchmarks}

\begin{table*}[t]
    \caption{Distribution-based metrics from \textit{de novo} generation of \num{10000} structures using models trained on the $MP$-20 dataset with the structural frameworks present in the $MP$-20 test dataset. 
    We report comparisons to other \textit{de novo} generation models: DiffCSP, FlowMM, and OMatG-EncDecODE.
    Best scores in each category are bolded and second-best are underlined.}
 \label{tab:dist_eval}
 \vspace{3pt}
  \resizebox{\textwidth}{!}{
  \centering
  \begin{tabular}
  {lcccccccccc}
    \toprule
    \multirow{2}{*}{\textbf{Method}} & \multirow{2}{*}{\textbf{Integration steps}} & \multicolumn{3}{c}{\textbf{Validity} (\% $\uparrow$)} & \multicolumn{5}{c}{\textbf{Property} ($\downarrow$)} \\
    \cmidrule(lr){3-5}\cmidrule(lr){6-10}
    &  & Structural & Composition & Combined & wdist ($\rho$) & wdist ($N$ary) & wdist $\pt{\langle CN \rangle_S}$ & wdist (elem) & wdist (sg) \\
    \midrule
    DiffCSP        & 1000 & \underline{99.91}          & 82.46         & 82.46          & 0.3133 & 0.3193 & 0.3916 & 1.6300 & 20.40 \\
    FlowMM         & 1000 & 92.26          & 76.87         & 76.86          & 1.0712 & 0.1130          & 0.6603 & 0.9255 & 13.02 \\
    OMatG-EncDecODE & 840 & 97.25          & 84.16        & 84.16          & \underline{0.1155} & \textbf{0.0553} & \textbf{0.1026} & \underline{0.8249} & 24.80 \\
    \midrule
    OMatG-FC-LinearODE & 610 & \textbf{100.0} & \textbf{87.38} & \textbf{87.38} & 0.6045   & \underline{0.0719}  & 0.1890 & \textbf{0.4038} & \textbf{4.413} \\
    OMatG-FC-LinearSDE & 620 & \textbf{100.0} & \underline{86.33} & \underline{86.32} & \textbf{0.0860} & 0.0856  & \underline{0.1163} & 0.8483 & \underline{5.888} \\
  \bottomrule
\end{tabular}
}
\end{table*}

\begin{table*}[ht]
\caption{Benchmarks for \textit{de novo} generation as provided by the LeMat-GenBench software \citep{betala_lematgenbench_2026} on \num{2500} generated structures. 
Models are trained and evaluated on the $MP$-20 dataset.
Stability is defined as having $ \Delta E_{\mathrm{hull}} \leq 0$ eV/atom above hull and metastability is defined as $0 < \Delta E_{\mathrm{hull}}  \leq 0.1$ eV/atom above hull. 
Uniqueness and novelty results are combined with (meta)stability assessments to produce (M)SUN rates.
For OMatG and OMatG-FC models, structures are subsampled from those in Tab.~\ref{tab:dist_eval} and benchmarks are computed; for random enumeration and ionic substitution the benchmarking is described in Appendix~\ref{app:MGB}; all other benchmarks from other models are reproduced from the LeMat-GenBench leaderboard.
Models marked with a $\dagger$ utilize pre-relaxation before assessment; comparisons are therefore made within each category (unrelaxed and relaxed).
Best scores in each category are bolded and second-best are underlined.
}
\label{tab:dng_eval}
\vspace{3pt}
 \centering
  \resizebox{\textwidth}{!}{
  \begin{tabular}
  {lccccccccc}
    \toprule
    \multirow{2}{*}{Method} 
    & Validity & Unique
    & Novelty & Stable & Metastable & S.U.N. & M.S.U.N. 
    & $\langle E \rangle/N$ 
    & RMSD \\
    & Rate (\%, $\uparrow$)
    & Rate (\%, $\uparrow$)
    & Rate (\%, $\uparrow$) & Rate (\%, $\uparrow$) & Rate (\%, $\uparrow$) & Rate (\%, $\uparrow$) & Rate (\%, $\uparrow$)
    & (eV, $\downarrow)$
    & (\text{\AA}, $\downarrow$) \\
    \midrule

    Random enumeration \cite{szymanski_2025_matgen_baselines} & 70.4 & 70.3 & 61.4 & 0.0 & 0.1 & 0.0 & 0.1 & 2.82 & 1.76\\

    Ionic substitution \cite{szymanski_2025_matgen_baselines, hautier_data_2011a} & 33.8 & 32.5 & 20.7 & 0.6 & 10.0 & \underline{0.3} & 4.0 & 0.53 & 0.64 \\

    ADiT \cite{joshi_allatom_2025} & 90.6 & 87.8
    & 26.0 & 0.4
    & 36.5 & 0.0
    & 1.0 & 0.33 & 0.38\\

    DiffCSP \cite{jiao_crystal_2023} & 95.7 &  \underline{94.8}
    & \textbf{66.2} & \underline{2.3}
    & 29.8 & 0.1
    & \underline{8.5} & 0.27 & 0.59\\

    DiffCSP++ \cite{jiao_space_2024} & 95.3 & \textbf{95.1} 
    & \underline{62.0} & 1.0
    & 26.4 & 0.2
    & 5.0 & 0.41 & 0.69\\

    SymmCD \cite{levy_symmcd_2024} & 73.4 & 73.0 
    & 47.0 & 1.4
    & 18.6 & 0.1
    & 2.4 & 0.88 & 0.87\\

    Chemeleon2 \cite{Park2025chemeleon2} & 95.2 & 88.1 
    & 71.6 & 0.0
    & 39.8 & 0.0 
    & \textbf{21.2} & \textbf{0.15} & 0.42\\
    
    OMatG-EncDecODE \cite{hoellmer_open_2025a} & 92.4 & 91.6
    & 55.1 & 0.4
    & 30.3 & 0.1
    & 3.6 & 0.31 & 0.61\\

    \midrule

    OMatG-FC-LinearODE & \underline{95.9} & 92.2
    & 30.4 & 2.0
    & \underline{49.8} & \textbf{0.4}
    & 5.9 & 0.27 & \underline{0.31} \\

    OMatG-FC-LinearSDE & \textbf{96.5} & 93.4
    & 34.9 & \textbf{3.6}
    & \textbf{56.4} & 0.2
    & 6.8 & \underline{0.17} & \textbf{0.30} \\

    \midrule \midrule

    Ionic substitution$^\dagger$ & 34.4 & 33.1 & 20.6 & 4.7 & 19.2 & 1.2 & 10.8 & 0.09 & 0.09 \\

    MatterGen$^\dagger$ \cite{zeni_generative_2025} & 95.7 & 95.1 
    & \textbf{70.5} & 2.0 
    & 33.4 & 0.2 
    & 15.0 & 0.18 & 0.39 \\
    
    PLaID++$^\dagger$ \cite{xu_plaid_2025} & 96.0 & 77.8 
    & 24.2 & 12.4 
    & \textbf{60.7} & 1.0 
    & 7.6 & \underline{0.08} & 0.13\\

    Crystalite$^\dagger$ \cite{veljkovic_crystalite_2026} & \textbf{97.2} & \textbf{95.8} 
    & \underline{53.2} & 12.7 & 51.6
    & \underline{1.5} & \textbf{22.6} & 0.09 & 0.13\\

    OMatG-EncDecODE$^\dagger$ & \underline{96.4} & \underline{95.2}
    & 51.2 & 11.6
    & 49.8 & 1.0
    & \underline{18.0} & 0.09 & \underline{0.08}\\
    \midrule

    OMatG-FC-LinearODE$^\dagger$ & \textbf{97.2} & 92.8
    & 28.9 & \textbf{18.4}
    & \underline{57.9} & \textbf{1.7}
    & 12.0 & \textbf{0.07} & \textbf{0.07} \\

    OMatG-FC-LinearSDE$^\dagger$ & \textbf{97.2} & 93.8
    & 33.6 & \underline{16.0}
    & 57.5 & 1.0 & 13.3
    & \textbf{0.07} & 0.16 \\
    
  \bottomrule
\end{tabular}
}
\end{table*}

\begin{table*}[ht]
\caption{Benchmarks of models trained and evaluated on the $MPTS$-52 dataset for \textit{de novo} generation.
LeMat-GenBench \citep{betala_lematgenbench_2026} is used to evaluate \num{2500} generated structures.
The column definitions are the same as those of Tab.~\ref{tab:dng_eval}.
Models marked with a $\dagger$ utilize pre-relaxation before assessment.
}
\label{tab:mpts52_dng_eval}
\vspace{3pt}
 \centering
  \resizebox{\textwidth}{!}{
  \begin{tabular}
  {lccccccccc}
    \toprule
    \multirow{2}{*}{Method} 
    & Validity & Unique
    & Novelty & Stable & Metastable & S.U.N. & M.S.U.N. 
    & $\langle E \rangle/N$ 
    & RMSD \\
    & Rate (\%, $\uparrow$)
    & Rate (\%, $\uparrow$)
    & Rate (\%, $\uparrow$) & Rate (\%, $\uparrow$) & Rate (\%, $\uparrow$) & Rate (\%, $\uparrow$) & Rate (\%, $\uparrow$)
    & (eV, $\downarrow)$
    & (\text{\AA}, $\downarrow$) \\
    \midrule

    SGEquiDiff \cite{chang_space_2025} & 88.1 & \underline{87.7} & \textbf{69.1} & 0.4 & 7.3 & 0.1 & 1.0 & 0.9145 & 1.1722 \\
    Crystalite \cite{veljkovic_crystalite_2026} & 86.4 & 85.0 & 45.1 & \textbf{4.1} & \textbf{38.5} & \textbf{0.5} & 6.0 & \textbf{0.2770} & \textbf{0.5036} \\
    \hline
    
    OMatG-FC-LinearODE & \underline{92.9} & 87.3 & 64.7 & \underline{2.0} & 30.5 & \underline{0.4} & \textbf{8.4} & 0.4908 & \underline{0.6431} \\
    OMatG-FC-LinearSDE & \textbf{95.6} & \textbf{90.8} & \underline{65.7} & 1.5 & \underline{31.4} & 0.3 & \underline{7.9} & \underline{0.3521} & 0.6466 \\
    
    \midrule \midrule
    
    SGEquiDiff$^\dagger$ & 93.5 & \underline{93.1} & \textbf{69.0} & 6.3 & 23.1 & 0.6 & 10.9 & 0.2688 & 0.3003 \\
    Crystalite$^\dagger$ & \textbf{96.5} & \textbf{95.0} & 54.1 & \textbf{16.9} & \textbf{45.2} & \underline{1.6} & 21.2 & \textbf{0.1079} & \textbf{0.1758} \\
    \hline
    
    OMatG-FC-LinearODE$^\dagger$ & 94.6 & 89.2 & \underline{65.0} & 13.4 & 38.8 & \textbf{2.0} & \underline{21.8} & 0.1364 & 0.1919 \\
    OMatG-FC-LinearSDE$^\dagger$ & \underline{96.4} & 91.8 & 64.8 & \underline{13.6} & \underline{40.1} & \textbf{2.0} & \textbf{22.1} & \underline{0.1333} & \underline{0.1900} \\
    
  \bottomrule
\end{tabular}
}
\end{table*}

\subsection{Distributional Properties}\label{sec:dist_evals}

To evaluate whether our model has learned the data distribution, we compare distributional properties of generated structures against the test set using Wasserstein distance. The properties we compare include mass density $\rho$, $N$-arity (number of unique elements), averaged coordination number by element $\langle CN \rangle_S$, element counts, and space groups.
We also report validity metrics for the generated structures, including structural validity (all bond lengths $>0.5$\AA), compositional validity as measured by the SMACT software (version 3.0), and the combined validity requiring both criteria to be satisfied.


\subsection{Stability, Uniqueness, and Novelty}\label{sec:SUN}

We use the benchmarking framework LeMat-GenBench \cite{betala_lematgenbench_2026} to compare OMatG-FC performance on standardized metrics against other recent generative models. 
This benchmarking suite has the advantage of a public leaderboard for ease of comparison and unified benchmarks, and uses three machine-learned interatomic potentials---MACE \cite{batatia_foundation_2025}, UMA \cite{wood_uma_2025}, and Orb-v3 \cite{rhodes_orbv3_2025}---as a surrogate for DFT relaxation to compute stability rate, the energy above hull, and the root-mean-squared-distance (RMSD with \AA{} units) between the generated and relaxed structures.
A validity metric screens structures based on a number of parameters measuring physical plausibility, including CIF readability, structural, density, and lattice parameter validity, symmetry, and charge neutrality.
Uniqueness and novelty are assessed using pymatgen's \texttt{StructureMatcher} as a fingerprint method with \texttt{stol=0.1}.
Generated structures are compared against the large LeMat-Bulk reference dataset\cite{siron_lematbulk_2025} which includes the Materials Project\cite{jain_commentary_2013}, Alexandria\cite{Cavignac2026Alexandria}, and OQMD\cite{kirklin_open_2015a}. 
Structures are considered unique if they are non-redundant within the generated set, and structures are novel if they are absent from the reference dataset.
Stability is defined in terms of the energy above the convex hull ($\Delta E_{\textrm{hull}}$), where stable structures satisfy $\Delta E_{\textrm{hull}} \leq 0$ eV/atom and metastable structures satisfy $0 < \Delta E_{\textrm{hull}} \leq 0.1$ eV/atom.
Combined, these metrics define S.U.N (stable, unique, and novel) and M.S.U.N. (metastable, unique, and novel) rates which are the union of the three properties for each structure. 
We follow the LeMat-GenBench leaderboard standard and compute these metrics on 2,500 structures sampled from our generated test set.

\section{Results}

We present distributional metrics for generated structures in Tab.~\ref{tab:dist_eval}. OMatG-FC performs competitively on most distributional benchmarks, with notable improvement on the space group Wasserstein distance.\footnote{In Tab.~\ref{tab:dist_eval}, the space group Wasserstein distances are evaluated \textit{via} space group identification with \texttt{spglib} \cite{togo_spglib_2024}. Model performance is evaluated using a variable precision protocol introduced by OMatG or by setting \texttt{symprec}=\SI{1e-2}{}. 
The results in Tab.~\ref{tab:dist_eval} reflect in the lowest wdist(sg) values for each model from the precision protocol choices. For each fixed choice of tolerance handling, OMatG-FC models still outperform the others.}
This improvement arises because fixed positions and lattice parameters yield accurate symmetries by construction, whereas non-constrained models produce positional noise that requires relaxed symmetry tolerances for space group identification. For the same reason, structural validity is 100\%. 

Materials discovery metrics from LeMat-GenBench are shown in Tab.~\ref{tab:dng_eval} for models trained and evaluated on the $MP$-20 dataset.
OMatG-FC achieves state-of-the-art performance for S.U.N. rates within both unrelaxed and relaxed generated structures, lowest average energy above hull $\langle E \rangle / N$, and RMSD between generated and relaxed structures.
This is expected as structural frameworks from the dataset guide generated crystals closer to stable atomic packings.
OMatG-FC models also achieve the highest validity scores, which here include structural validity but also the metrics described in Sec.~\ref{sec:SUN}.
Finally, we demonstrate the highest stability rates and metastability rates, which compensate for lower novelty rates (similarly to the PLaID++ model \cite{xu_plaid_2025}).\looseness=-1

We additionally evaluate our model and two others, Crystalite \cite{veljkovic_crystalite_2026} and SGEquiDiff \cite{chang_space_2025}, on the $MPTS$-52 dataset, with benchmarking results shown in Tab.~\ref{tab:mpts52_dng_eval}.\footnote{We trained and sampled from a Crystalite model and sampled from an existing checkpoint for SGEquiDiff.}
OMatG-FC outperforms both in (M.)S.U.N. rates for relaxed structures.
For unrelaxed structures, OMatG-FC and Crystalite achieve comparable S.U.N. rates while OMatG-FC outperforms in M.S.U.N. rate. 
Notably, OMatG-FC's novelty rate is significantly higher on $MPTS$-52 than on $MP$-20.
We posit that this is due to the combinatorially larger structural space of $MPTS$-52---with more sites in the unit cell available to decorate with atomic species---allows the model to explore much more outside of the training dataset.


\subsection{Novel Prototypes}

\begin{figure}[t]
   \centering
   \includegraphics[width=\columnwidth]{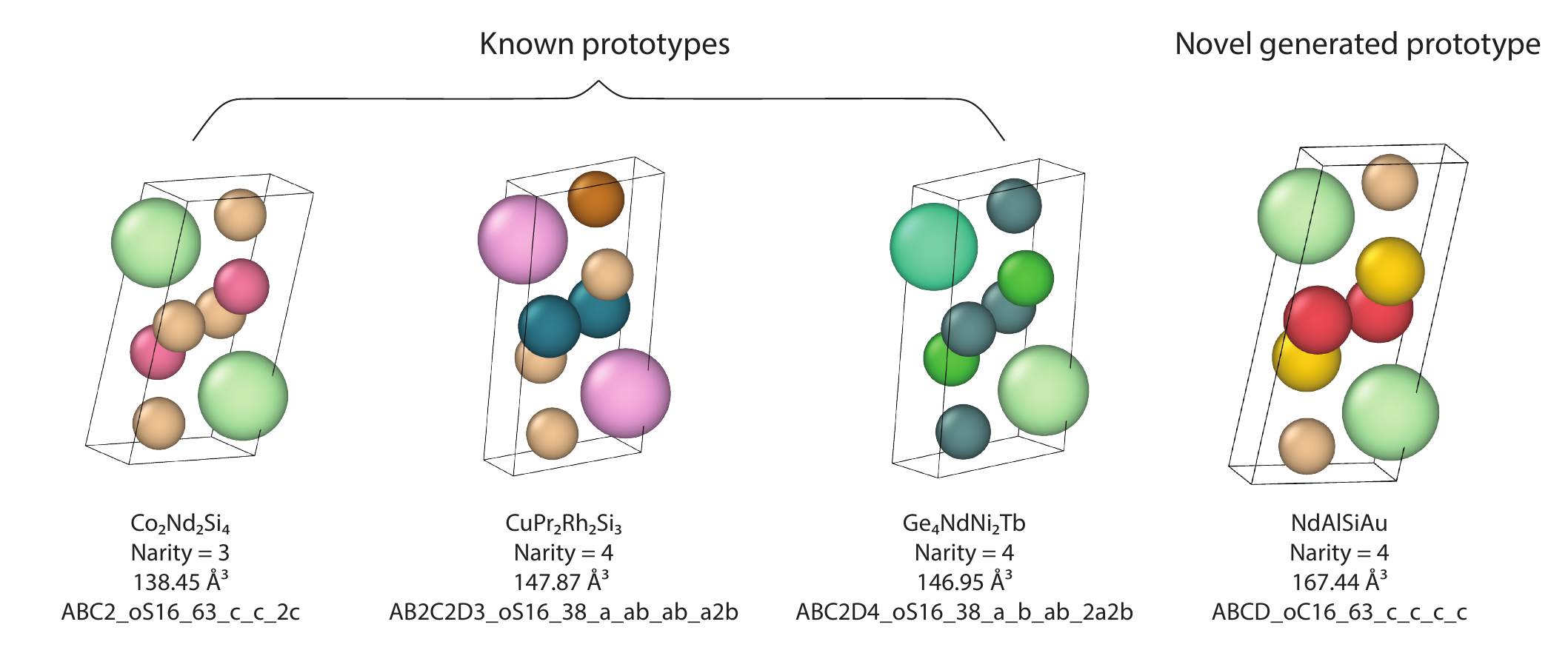}
   \caption{Three examples of known materials from the $MP$-20 dataset and a S.U.N. structure sampled using OMG-FC. All four materials share the same underlying structural framework but differ by prototype. 
   Labels denote the chemical composition, $N$-arity, volume, and AFLOW prototype label. 
   The sampled NdAlSiAu structure represents an unseen prototype for the model.
   }
   \label{fig:prototype}
\end{figure}

We examine the propensity of our method to discover prototype structures when conditioned on a framework. From the OMatG-FC-LinearODE model without pre-relaxation, we identify \ce{NdAlSiAu} as a stable structure occupying a prototype absent from the AFLOW Prototype Encyclopedia.
Using \texttt{StructureMatcher} with \texttt{stol}$=0.1$, we identified 124 materials in the $MP$20-dataset with the most similar structural frameworks as NdAlSiAu.
We find that there are only three distinct prototypes among them: 121 have $N$-arity $=3$ and share the same prototype represented by Co$_2$Nd$_2$Si$_4$ in Fig.~\ref{fig:prototype}, and the remaining three structures each with $N$-arity $=4$ showcase two additional prototypes.
OMatG-FC is able to sample a fourth unseen prototype.
This highlights a particular advance over prototype-constrained methods---our framework conditioning, applied within a stochastic interpolant model, lifts that restriction.

\section{Discussion and Conclusion}

We introduce a new type of generative model, OMatG-FC, which allows for the exploration of novel chemical compositions conditional on structural frameworks.
While counterintuitive for the prospect of discovering novel materials, we show that our method is a state-of-the-art generative model for materials discovery.
This is accomplished through simplification of the learning task through the conditional prior, as well as the inductive bias imposed by the structural framework constraint which pushes the learned distribution towards crystals that are closer to the convex hull (\textit{i.e.}, more stable).
Ultimately, OMatG-FC paves the way for discovery \textit{via} structural frameworks---limited in our study to those from the dataset, but in practice applicable to any theorized point pattern.

\section*{Acknowledgments}
The authors would like to thank Shenglong Wang at NYU IT HPC and Gregory Wolfe for supporting this work.
The authors acknowledge funding from NSF Grant OAC-2311632.
P.\ H.\ and S.\ M.\ also acknowledge support from the Simons Center for Computational Physical Chemistry (Simons Foundation grant 839534, MT).
The authors gratefully acknowledge the computational resources, consultation support, and core personnel that have contributed to the research results reported in this publication, provided by the IT High Performance Computing at New York University and NYU Langone High Performance Computing.

\newpage
\bibliography{matdisc}

@article{Cavignac2026Alexandria,
  author  = {Cavignac, Th{\'e}o and Schmidt, Jonathan and De Breuck, Pierre-Paul and Loew, Antoine
             and Cerqueira, Tiago F. T. and Wang, Hai-Chen and Bochkarev, Anton
             and Lysogorskiy, Yury and Romero, Aldo H. and Drautz, Ralf and Botti, Silvana
             and Marques, Miguel A. L.},
  title   = {{AI}-Driven expansion and application of the {A}lexandria database},
  journal = {Journal of Physics: Materials},
  year    = {2026},
  volume  = {9},
  number  = {2},
  pages   = {025014},
  doi     = {10.1088/2515-7639/ae6620},
  url     = {https://doi.org/10.1088/2515-7639/ae6620}
}

@misc{siron_lematbulk_2025,
  title = {{{LeMat-Bulk}}: Aggregating, and de-Duplicating Quantum Chemistry Materials Databases},
  shorttitle = {{{LeMat-Bulk}}},
  author = {Siron, Martin and Djafar, Inel and Ramlaoui, Ali and du Fayette, Etienne and Rossello, Amandine and Fako, Edvin and McDermott, Matthew and Therrien, Felix and {Barroso-Luque}, Luis and Cipcigan, Flaviu and Schwaller, Philippe and Wolf, Thomas and Duval, Alexandre},
  year = 2025,
  month = nov,
  number = {arXiv:2511.05178},
  eprint = {2511.05178},
  primaryclass = {cond-mat.mtrl-sci},
  publisher = {arXiv},
  doi = {10.48550/arXiv.2511.05178},
  urldate = {2026-10-28},
  archiveprefix = {arXiv}
}

@misc{veljkovic_crystalite_2026,
  title = {Crystalite: {{A Lightweight Transformer}} for {{Efficient Crystal Modeling}}},
  shorttitle = {Crystalite},
  author = {Veljkovi{\'c}, Tin Had{\v z}i and Rosenthal, Joshua and Lon{\v c}ari{\'c}, Ivor and van de Meent, Jan-Willem},
  year = 2026,
  month = apr,
  number = {arXiv:2604.02270},
  eprint = {2604.02270},
  primaryclass = {cs},
  publisher = {arXiv},
  doi = {10.48550/arXiv.2604.02270},
  urldate = {2026-04-03},
  archiveprefix = {arXiv}
}

@article{Park2025chemeleon2,
  title={Guiding Generative Models to Uncover Diverse and Novel Crystals via Reinforcement Learning},
  author={Hyunsoo Park and Aron Walsh},
  year={2025},
  url={https://arxiv.org/abs/2511.07158}
}

@article{szymanski_2025_matgen_baselines,
    title={Establishing baselines for generative discovery of inorganic crystals},
    DOI={10.48550/arXiv.2501.02144},
    journal={arXiv},
    author={Szymanski, Nathan J. and Bartel, Christopher J.},
    year={2025}
}

@article{jakob_learning_2026,
	title = {Learning {Crystallographic} {Disorder}: {Bridging} {Prediction} and {Experiment} in {Materials} {Discovery}},
	volume = {38},
	issn = {0935-9648, 1521-4095},
	shorttitle = {Learning {Crystallographic} {Disorder}},
	url = {https://advanced.onlinelibrary.wiley.com/doi/10.1002/adma.202514226},
	doi = {10.1002/adma.202514226},
	language = {en},
	number = {5},
	urldate = {2026-02-01},
	journal = {Advanced Materials},
	author = {Jakob, Konstantin S. and Walsh, Aron and Reuter, Karsten and Margraf, Johannes T.},
	month = jan,
	year = {2026},
	pages = {e14226},
}

@article{antypov_classification_2025,
	title = {Classification and statistical analysis of structural disorder in crystalline materials},
	volume = {58},
	issn = {1600-5767},
	url = {https://journals.iucr.org/paper?S1600576725003000},
	doi = {10.1107/S1600576725003000},
	language = {en},
	number = {3},
	urldate = {2026-02-01},
	journal = {Journal of Applied Crystallography},
	author = {Antypov, Dmytro and Collins, Chris M. and Dyer, Matthew S. and Claridge, John B. and Rosseinsky, Matthew J.},
	month = jun,
	year = {2025},
	pages = {659--677},
}

@misc{akiba_optuna_2019,
  title = {Optuna: {{A Next-generation Hyperparameter Optimization Framework}}},
  shorttitle = {Optuna},
  author = {Akiba, Takuya and Sano, Shotaro and Yanase, Toshihiko and Ohta, Takeru and Koyama, Masanori},
  year = {2019},
  month = jul,
  number = {arXiv:1907.10902},
  eprint = {1907.10902},
  primaryclass = {cs},
  publisher = {arXiv},
  doi = {10.48550/arXiv.1907.10902},
  url = {http://arxiv.org/abs/1907.10902},
  urldate = {2026-01-29},
  archiveprefix = {arXiv}
}

@article{liaw_tune_2018,
    title={Tune: A Research Platform for Distributed Model Selection and Training},
    author={Liaw, Richard and Liang, Eric and Nishihara, Robert
            and Moritz, Philipp and Gonzalez, Joseph E and Stoica, Ion},
    journal={arXiv preprint arXiv:1807.05118},
    year={2018}
}

@misc{xu_plaid_2025,
  title = {{{PLaID}}++: {{A Preference Aligned Language Model}} for {{Targeted Inorganic Materials Design}}},
  shorttitle = {{{PLaID}}++},
  author = {Xu, Andy and Desai, Rohan and Wang, Larry and Hope, Gabriel and Ritz, Ethan},
  year = {2025},
  month = sep,
  number = {arXiv:2509.07150},
  eprint = {2509.07150},
  primaryclass = {cs},
  publisher = {arXiv},
  doi = {10.48550/arXiv.2509.07150},
  url = {http://arxiv.org/abs/2509.07150},
  urldate = {2026-01-29},
  archiveprefix = {arXiv}
}

@misc{prakash_guided_2025,
  title = {Guided {{Diffusion}} for the {{Discovery}} of {{New Superconductors}}},
  author = {Prakash, Pawan and Gibson, Jason B. and Li, Zhongwei and Gianluca, Gabriele Di and Esquivel, Juan and Fuemmeler, Eric and Geisler, Benjamin and Kim, Jung Soo and Roitberg, Adrian and Tadmor, Ellad B. and Liu, Mingjie and Martiniani, Stefano and Stewart, Gregory R. and Hamlin, James J. and Hirschfeld, Peter J. and Hennig, Richard G.},
  year = {2025},
  month = sep,
  number = {arXiv:2509.25186},
  eprint = {2509.25186},
  primaryclass = {cond-mat},
  publisher = {arXiv},
  doi = {10.48550/arXiv.2509.25186},
  url = {http://arxiv.org/abs/2509.25186},
  urldate = {2026-01-29},
  archiveprefix = {arXiv}
}

@misc{rhodes_orbv3_2025,
  title = {Orb-v3: Atomistic Simulation at Scale},
  shorttitle = {Orb-V3},
  author = {Rhodes, Benjamin and Vandenhaute, Sander and {\v S}imkus, Vaidotas and Gin, James and Godwin, Jonathan and Duignan, Tim and Neumann, Mark},
  year = {2025},
  month = apr,
  number = {arXiv:2504.06231},
  eprint = {2504.06231},
  primaryclass = {cond-mat},
  publisher = {arXiv},
  doi = {10.48550/arXiv.2504.06231},
  url = {http://arxiv.org/abs/2504.06231},
  urldate = {2026-01-29},
  archiveprefix = {arXiv}
}

@misc{bone_discovery_2025,
  title = {Discovery and Recovery of Crystalline Materials with Property-Conditioned Transformers},
  author = {Bone, Cyprien and Walker, Matthew and Leng, Kuangdai and Antunes, Luis M. and {Grau-Crespo}, Ricardo and Aligayev, Amil and Dominguez, Javier and Butler, Keith T.},
  year = {2025},
  month = nov,
  number = {arXiv:2511.21299},
  eprint = {2511.21299},
  primaryclass = {cond-mat},
  publisher = {arXiv},
  doi = {10.48550/arXiv.2511.21299},
  url = {http://arxiv.org/abs/2511.21299},
  urldate = {2026-01-28},
  archiveprefix = {arXiv}
}

@misc{metni_generative_2025,
  title = {Generative Models for Crystalline Materials},
  author = {Metni, Houssam and Ruple, Laura and Walters, Lauren N. and Torresi, Luca and Teufel, Jonas and Schopmans, Henrik and {\"O}streicher, Jona and Zhang, Yumeng and Neubert, Marlen and Koide, Yuri and Steiner, Kevin and Link, Paul and B{\"a}r, Lukas and Petrova, Mariana and Ceder, Gerbrand and Friederich, Pascal},
  year = {2025},
  month = nov,
  number = {arXiv:2511.22652},
  eprint = {2511.22652},
  primaryclass = {cond-mat},
  publisher = {arXiv},
  doi = {10.48550/arXiv.2511.22652},
  url = {http://arxiv.org/abs/2511.22652},
  urldate = {2026-01-28},
  archiveprefix = {arXiv}
}

@article{nomura_room-temperature_2004,
	title = {Room-temperature fabrication of transparent flexible thin-film transistors using amorphous oxide semiconductors},
	volume = {432},
	copyright = {http://www.springer.com/tdm},
	issn = {0028-0836, 1476-4687},
	url = {https://www.nature.com/articles/nature03090},
	doi = {10.1038/nature03090},
	language = {en},
	number = {7016},
	urldate = {2026-01-28},
	journal = {Nature},
	author = {Nomura, Kenji and Ohta, Hiromichi and Takagi, Akihiro and Kamiya, Toshio and Hirano, Masahiro and Hosono, Hideo},
	month = nov,
	year = {2004},
	pages = {488--492},
}

@article{kamaya_lithium_2011,
	title = {A lithium superionic conductor},
	volume = {10},
	issn = {1476-1122, 1476-4660},
	url = {https://www.nature.com/articles/nmat3066},
	doi = {10.1038/nmat3066},
	language = {en},
	number = {9},
	urldate = {2026-01-28},
	journal = {Nature Materials},
	author = {Kamaya, Noriaki and Homma, Kenji and Yamakawa, Yuichiro and Hirayama, Masaaki and Kanno, Ryoji and Yonemura, Masao and Kamiyama, Takashi and Kato, Yuki and Hama, Shigenori and Kawamoto, Koji and Mitsui, Akio},
	month = sep,
	year = {2011},
	pages = {682--686},
}

@article{jang_structurally_2024,
	title = {Structurally robust lithium-rich layered oxides for high-energy and long-lasting cathodes},
	volume = {15},
	issn = {2041-1723},
	url = {https://www.nature.com/articles/s41467-024-45490-x},
	doi = {10.1038/s41467-024-45490-x},
	language = {en},
	number = {1},
	urldate = {2026-01-28},
	journal = {Nature Communications},
	author = {Jang, Ho-Young and Eum, Donggun and Cho, Jiung and Lim, Jun and Lee, Yeji and Song, Jun-Hyuk and Park, Hyeokjun and Kim, Byunghoon and Kim, Do-Hoon and Cho, Sung-Pyo and Jo, Sugeun and Heo, Jae Hoon and Lee, Sunyoung and Lim, Jongwoo and Kang, Kisuk},
	month = feb,
	year = {2024},
	pages = {1288},
}

@article{cao_space_2025,
  title = {Space {{Group Informed Transformer}} for {{Crystalline Materials Generation}}},
  author = {Cao, Zhendong and Luo, Xiaoshan and Lv, Jian and Wang, Lei},
  year = {2025},
  month = nov,
  journal = {Science Bulletin},
  volume = {70},
  number = {21},
  eprint = {2403.15734},
  primaryclass = {cond-mat},
  pages = {3522--3533},
  issn = {20959273},
  doi = {10.1016/j.scib.2025.09.035},
  url = {http://arxiv.org/abs/2403.15734},
  urldate = {2026-01-28},
  archiveprefix = {arXiv}
}

@article{vu_towards_2023,
	title = {Towards understanding structure–property relations in materials with interpretable deep learning},
	volume = {9},
	issn = {2057-3960},
	url = {https://www.nature.com/articles/s41524-023-01163-9},
	doi = {10.1038/s41524-023-01163-9},
	language = {en},
	number = {1},
	urldate = {2026-01-28},
	journal = {npj Computational Materials},
	author = {Vu, Tien-Sinh and Ha, Minh-Quyet and Nguyen, Duong-Nguyen and Nguyen, Viet-Cuong and Abe, Yukihiro and Tran, Truyen and Tran, Huan and Kino, Hiori and Miyake, Takashi and Tsuda, Koji and Dam, Hieu-Chi},
	month = dec,
	year = {2023},
	pages = {215},
}

@article{de_breuck_generative_2025,
	title = {Generative {AI} for crystal structures: a review},
	volume = {11},
	issn = {2057-3960},
	shorttitle = {Generative {AI} for crystal structures},
	url = {https://www.nature.com/articles/s41524-025-01881-2},
	doi = {10.1038/s41524-025-01881-2},
	language = {en},
	number = {1},
	urldate = {2026-01-28},
	journal = {npj Computational Materials},
	author = {De Breuck, Pierre-Paul and Wang, Hai-Chen and Rignanese, Gian-Marco and Botti, Silvana and Marques, Miguel A. L.},
	month = dec,
	year = {2025},
	pages = {370},
}

@article{eckert_aflow_2024a,
  title = {The {{AFLOW Library}} of {{Crystallographic Prototypes}}: {{Part}} 4},
  shorttitle = {The {{AFLOW Library}} of {{Crystallographic Prototypes}}},
  author = {Eckert, Hagen and Divilov, Simon and Mehl, Michael J. and Hicks, David and Zettel, Adam C. and Esters, Marco and Campilongo, Xiomara and Curtarolo, Stefano},
  year = {2024},
  month = may,
  journal = {Computational Materials Science},
  volume = {240},
  eprint = {2401.06875},
  primaryclass = {cond-mat},
  pages = {112988},
  issn = {09270256},
  doi = {10.1016/j.commatsci.2024.112988},
  url = {http://arxiv.org/abs/2401.06875},
  urldate = {2026-01-28},
  archiveprefix = {arXiv}
}

@article{hicks_aflowxtalfinder_2021,
  title = {{{AFLOW-XtalFinder}}: A Reliable Choice to Identify Crystalline Prototypes},
  shorttitle = {{{AFLOW-XtalFinder}}},
  author = {Hicks, David and Toher, Cormac and Ford, Denise C. and Rose, Frisco and Santo, Carlo De and Levy, Ohad and Mehl, Michael J. and Curtarolo, Stefano},
  year = {2021},
  month = feb,
  journal = {npj Computational Materials},
  volume = {7},
  number = {1},
  pages = {30},
  publisher = {Nature Publishing Group},
  issn = {2057-3960},
  doi = {10.1038/s41524-020-00483-4},
  url = {https://www.nature.com/articles/s41524-020-00483-4},
  urldate = {2026-01-28},
  copyright = {2021 The Author(s)},
  langid = {english}
}

@book{hume-rothery_structure_1939,
  title = {The {{Structure}} of {{Metals}} and {{Alloys}}},
  author = {{Hume-Rothery}, William},
  year = {1939},
  publisher = {Chemical Publishing Company},
  googlebooks = {8A1NAAAAYAAJ},
  langid = {english}
}

@misc{betala_lematgenbench_2026,
  title = {{{LeMat-GenBench}}: {{A Unified Evaluation Framework}} for {{Crystal Generative Models}}},
  shorttitle = {{{LeMat-GenBench}}},
  author = {Betala, Siddharth and Gleason, Samuel P. and Ramlaoui, Ali and Xu, Andy and Channing, Georgia and Levy, Daniel and Fourrier, Cl{\'e}mentine and Kazeev, Nikita and Joshi, Chaitanya K. and Kaba, S{\'e}kou-Oumar and Therrien, F{\'e}lix and {Hernandez-Garcia}, Alex and Mercado, Roc{\'i}o and Krishnan, N. M. Anoop and Duval, Alexandre},
  year = {2026},
  month = jan,
  number = {arXiv:2512.04562},
  eprint = {2512.04562},
  primaryclass = {cs},
  publisher = {arXiv},
  doi = {10.48550/arXiv.2512.04562},
  url = {http://arxiv.org/abs/2512.04562},
  urldate = {2026-01-27},
  archiveprefix = {arXiv}
}

@inproceedings{
martirossyan_all_2025,
title={All that structure matches does not glitter},
author={Maya Martirossyan and Thomas Egg and Philipp H{\"o}llmer and George Karypis and Mark Transtrum and Adrian Roitberg and Mingjie Liu and Richard Hennig and Ellad B. Tadmor and Stefano Martiniani},
booktitle={The Thirty-ninth Annual Conference on Neural Information Processing Systems Datasets and Benchmarks Track},
year={2025},
url={https://openreview.net/forum?id=ig9ujp50D4}
}

@article{okabe_structural_2025a,
  title = {Structural Constraint Integration in a Generative Model for the Discovery of Quantum Materials},
  author = {Okabe, Ryotaro and Cheng, Mouyang and Chotrattanapituk, Abhijatmedhi and Mandal, Manasi and Mak, Kiran and C{\'o}rdova Carrizales, Denisse and Hung, Nguyen Tuan and Fu, Xiang and Han, Bowen and Wang, Yao and Xie, Weiwei and Cava, Robert J. and Jaakkola, Tommi S. and Cheng, Yongqiang and Li, Mingda},
  year = {2025},
  month = sep,
  journal = {Nature Materials},
  pages = {1--8},
  publisher = {Nature Publishing Group},
  issn = {1476-4660},
  doi = {10.1038/s41563-025-02355-y},
  url = {https://www.nature.com/articles/s41563-025-02355-y},
  urldate = {2026-01-19},
  copyright = {2025 The Author(s), under exclusive licence to Springer Nature Limited},
  langid = {english}
}

@misc{albergo_building_2023,
  title = {Building {{Normalizing Flows}} with {{Stochastic Interpolants}}},
  author = {Albergo, Michael S. and {Vanden-Eijnden}, Eric},
  year = {2023},
  month = mar,
  number = {arXiv:2209.15571},
  eprint = {2209.15571},
  primaryclass = {cs, stat},
  publisher = {arXiv},
  doi = {10.48550/arXiv.2209.15571},
  url = {http://arxiv.org/abs/2209.15571},
  urldate = {2024-09-23},
  archiveprefix = {arXiv}
}

@misc{albergo_stochastic_2023,
  title = {Stochastic {{Interpolants}}: {{A Unifying Framework}} for {{Flows}} and {{Diffusions}}},
  shorttitle = {Stochastic {{Interpolants}}},
  author = {Albergo, Michael S. and Boffi, Nicholas M. and {Vanden-Eijnden}, Eric},
  year = {2023},
  month = nov,
  number = {arXiv:2303.08797},
  eprint = {2303.08797},
  primaryclass = {cond-mat},
  publisher = {arXiv},
  doi = {10.48550/arXiv.2303.08797},
  url = {http://arxiv.org/abs/2303.08797},
  urldate = {2024-09-23},
  archiveprefix = {arXiv}
}

@article{baird_matbenchgenmetrics_2024,
  title = {Matbench-Genmetrics: {{A Python}} Library for Benchmarking Crystal Structure Generative Models Using Time-Based Splits of {{Materials Project}} Structures},
  shorttitle = {Matbench-Genmetrics},
  author = {Baird, Sterling G. and Sayeed, Hasan M. and Montoya, Joseph and Sparks, Taylor D.},
  year = {2024},
  month = may,
  journal = {Journal of Open Source Software},
  volume = {9},
  number = {97},
  pages = {5618},
  issn = {2475-9066},
  doi = {10.21105/joss.05618},
  url = {https://joss.theoj.org/papers/10.21105/joss.05618},
  urldate = {2025-01-29},
  langid = {english}
}

@article{batatia_foundation_2025,
  title = {A Foundation Model for Atomistic Materials Chemistry},
  author = {Batatia, Ilyes and Benner, Philipp and Chiang, Yuan and Elena, Alin M. and Kov{\'a}cs, D{\'a}vid P. and Riebesell, Janosh and Advincula, Xavier R. and Asta, Mark and Avaylon, Matthew and Baldwin, William J. and Berger, Fabian and Bernstein, Noam and Bhowmik, Arghya and Bigi, Filippo and Blau, Samuel M. and C{\u a}rare, Vlad and Ceriotti, Michele and Chong, Sanggyu and Darby, James P. and De, Sandip and Della Pia, Flaviano and Deringer, Volker L. and Elijo{\v s}ius, Rokas and {El-Machachi}, Zakariya and Fako, Edvin and Falcioni, Fabio and Ferrari, Andrea C. and Gardner, John L. A. and Gawkowski, Miko{\l}aj J. and {Genreith-Schriever}, Annalena and George, Janine and Goodall, Rhys E. A. and Grandel, Jonas and Grey, Clare P. and Grigorev, Petr and Han, Shuang and Handley, Will and Heenen, Hendrik H. and Hermansson, Kersti and Ho, Cheuk Hin and Hofmann, Stephan and Holm, Christian and Jaafar, Jad and Jakob, Konstantin S. and Jung, Hyunwook and Kapil, Venkat and Kaplan, Aaron D. and Karimitari, Nima and Kermode, James R. and Kourtis, Panagiotis and Kroupa, Namu and Kullgren, Jolla and Kuner, Matthew C. and Kuryla, Domantas and Liepuoniute, Guoda and Lin, Chen and Margraf, Johannes T. and Magd{\u a}u, Ioan-Bogdan and Michaelides, Angelos and Moore, J. Harry and Naik, Aakash A. and Niblett, Samuel P. and Norwood, Sam Walton and O'Neill, Niamh and Ortner, Christoph and Persson, Kristin A. and Reuter, Karsten and Rosen, Andrew S. and Rosset, Louise A. M. and Schaaf, Lars L. and Schran, Christoph and Shi, Benjamin X. and Sivonxay, Eric and Stenczel, Tam{\'a}s K. and Sutton, Christopher and Svahn, Viktor and Swinburne, Thomas D. and Tilly, Jules and {van der Oord}, Cas and Vargas, Santiago and {Varga-Umbrich}, Eszter and Vegge, Tejs and Vondr{\'a}k, Martin and Wang, Yangshuai and Witt, William C. and Wolf, Thomas and Zills, Fabian and Cs{\'a}nyi, G{\'a}bor},
  year = {2025},
  month = nov,
  journal = {J. Chem. Phys.},
  volume = {163},
  number = {18},
  pages = {184110},
  issn = {0021-9606},
  doi = {10.1063/5.0297006},
  url = {https://doi.org/10.1063/5.0297006},
  urldate = {2025-11-19}
}

@article{boeri_2021_2022,
  title = {The 2021 Room-Temperature Superconductivity Roadmap},
  author = {Boeri, Lilia and Hennig, Richard and Hirschfeld, Peter and Profeta, Gianni and Sanna, Antonio and Zurek, Eva and Pickett, Warren E and Amsler, Maximilian and Dias, Ranga and Eremets, Mikhail I and Heil, Christoph and Hemley, Russell J and Liu, Hanyu and Ma, Yanming and Pierleoni, Carlo and Kolmogorov, Aleksey N and Rybin, Nikita and Novoselov, Dmitry and Anisimov, Vladimir and Oganov, Artem R and Pickard, Chris J and Bi, Tiange and Arita, Ryotaro and Errea, Ion and Pellegrini, Camilla and Requist, Ryan and Gross, E K U and Margine, Elena Roxana and Xie, Stephen R and Quan, Yundi and Hire, Ajinkya and Fanfarillo, Laura and Stewart, G R and Hamlin, J J and Stanev, Valentin and Gonnelli, Renato S and Piatti, Erik and Romanin, Davide and Daghero, Dario and Valenti, Roser},
  year = {2022},
  month = mar,
  journal = {J. Phys.: Condens. Matter},
  volume = {34},
  number = {18},
  pages = {183002},
  publisher = {IOP Publishing},
  issn = {0953-8984},
  doi = {10.1088/1361-648X/ac2864},
  url = {https://dx.doi.org/10.1088/1361-648X/ac2864},
  urldate = {2025-01-27},
  langid = {english}
}

@misc{bose_se3stochastic_2024,
  title = {{{SE}}(3)-{{Stochastic Flow Matching}} for {{Protein Backbone Generation}}},
  author = {Bose, Avishek Joey and {Akhound-Sadegh}, Tara and Huguet, Guillaume and Fatras, Kilian and {Rector-Brooks}, Jarrid and Liu, Cheng-Hao and Nica, Andrei Cristian and Korablyov, Maksym and Bronstein, Michael and Tong, Alexander},
  year = {2024},
  month = apr,
  number = {arXiv:2310.02391},
  eprint = {2310.02391},
  publisher = {arXiv},
  doi = {10.48550/arXiv.2310.02391},
  url = {http://arxiv.org/abs/2310.02391},
  urldate = {2024-11-20},
  archiveprefix = {arXiv}
}

@misc{campbell_generative_2024,
  title = {Generative {{Flows}} on {{Discrete State-Spaces}}: {{Enabling Multimodal Flows}} with {{Applications}} to {{Protein Co-Design}}},
  shorttitle = {Generative {{Flows}} on {{Discrete State-Spaces}}},
  author = {Campbell, Andrew and Yim, Jason and Barzilay, Regina and Rainforth, Tom and Jaakkola, Tommi},
  year = {2024},
  month = jun,
  number = {arXiv:2402.04997},
  eprint = {2402.04997},
  primaryclass = {cs, q-bio, stat},
  publisher = {arXiv},
  doi = {10.48550/arXiv.2402.04997},
  url = {http://arxiv.org/abs/2402.04997},
  urldate = {2024-09-24},
  archiveprefix = {arXiv}
}

@misc{chang_space_2025,
  title = {Space {{Group Equivariant Crystal Diffusion}}},
  author = {Chang, Rees and Pak, Angela and Guerra, Alex and Zhan, Ni and Richardson, Nick and Ertekin, Elif and Adams, Ryan P.},
  year = {2025},
  month = may,
  number = {arXiv:2505.10994},
  eprint = {2505.10994},
  primaryclass = {cond-mat},
  publisher = {arXiv},
  doi = {10.48550/arXiv.2505.10994},
  url = {http://arxiv.org/abs/2505.10994},
  urldate = {2025-06-25},
  archiveprefix = {arXiv}
}

@article{cheetham_artificial_2024,
  title = {Artificial {{Intelligence Driving Materials Discovery}}? {{Perspective}} on the {{Article}}: {{Scaling Deep Learning}} for {{Materials Discovery}}},
  shorttitle = {Artificial {{Intelligence Driving Materials Discovery}}?},
  author = {Cheetham, Anthony K. and Seshadri, Ram},
  year = {2024},
  month = apr,
  journal = {Chem. Mater.},
  volume = {36},
  number = {8},
  pages = {3490--3495},
  publisher = {American Chemical Society},
  issn = {0897-4756},
  doi = {10.1021/acs.chemmater.4c00643},
  url = {https://doi.org/10.1021/acs.chemmater.4c00643},
  urldate = {2024-12-10}
}

@article{curtarolo_highthroughput_2013,
  title = {The High-Throughput Highway to Computational Materials Design},
  author = {Curtarolo, Stefano and Hart, Gus L. W. and Nardelli, Marco Buongiorno and Mingo, Natalio and Sanvito, Stefano and Levy, Ohad},
  year = {2013},
  month = mar,
  journal = {Nat. Mater.},
  volume = {12},
  number = {3},
  pages = {191--201},
  publisher = {Nature Publishing Group},
  issn = {1476-4660},
  doi = {10.1038/nmat3568},
  url = {https://www.nature.com/articles/nmat3568},
  urldate = {2024-01-12},
  copyright = {2013 Springer Nature Limited},
  langid = {english}
}

@misc{gat_discrete_2024,
  title = {Discrete {{Flow Matching}}},
  author = {Gat, Itai and Remez, Tal and Shaul, Neta and Kreuk, Felix and Chen, Ricky T. Q. and Synnaeve, Gabriel and Adi, Yossi and Lipman, Yaron},
  year = {2024},
  month = jul,
  number = {arXiv:2407.15595},
  eprint = {2407.15595},
  primaryclass = {cs},
  publisher = {arXiv},
  doi = {10.48550/arXiv.2407.15595},
  url = {http://arxiv.org/abs/2407.15595},
  urldate = {2024-09-23},
  archiveprefix = {arXiv}
}

@article{gludovatz_exceptional_2016,
  title = {Exceptional Damage-Tolerance of a Medium-Entropy Alloy {{CrCoNi}} at Cryogenic Temperatures},
  author = {Gludovatz, Bernd and Hohenwarter, Anton and Thurston, Keli V. S. and Bei, Hongbin and Wu, Zhenggang and George, Easo P. and Ritchie, Robert O.},
  year = {2016},
  month = feb,
  journal = {Nat. Commun.},
  volume = {7},
  number = {1},
  pages = {10602},
  publisher = {Nature Publishing Group},
  issn = {2041-1723},
  doi = {10.1038/ncomms10602},
  url = {https://www.nature.com/articles/ncomms10602},
  urldate = {2024-01-18},
  copyright = {2016 The Author(s)},
  langid = {english}
}

@article{hautier_data_2011a,
  title = {Data {{Mined Ionic Substitutions}} for the {{Discovery}} of {{New Compounds}}},
  author = {Hautier, Geoffroy and Fischer, Chris and Ehrlacher, Virginie and Jain, Anubhav and Ceder, Gerbrand},
  year = {2011},
  month = jan,
  journal = {Inorg. Chem.},
  volume = {50},
  number = {2},
  pages = {656--663},
  publisher = {American Chemical Society},
  issn = {0020-1669},
  doi = {10.1021/ic102031h},
  url = {https://doi.org/10.1021/ic102031h},
  urldate = {2025-07-01}
}

@inproceedings{
    hoellmer_open_2025a,
    title={{Open Materials Generation with Stochastic Interpolants}},
    author={Philipp H{\"o}llmer and Thomas Egg and Maya Martirossyan and Eric Fuemmeler and Zeren Shui and Amit Gupta and Pawan Prakash and Adrian Roitberg and Mingjie Liu and George Karypis and Mark Transtrum and Richard Hennig and Ellad B. Tadmor and Stefano Martiniani},
    booktitle={Forty-second International Conference on Machine Learning},
    year={2025},
    url={https://openreview.net/forum?id=gHGrzxFujU}
}

@article{jain_commentary_2013,
  title = {Commentary: {{The Materials Project}}: {{A}} Materials Genome Approach to Accelerating Materials Innovation},
  shorttitle = {Commentary},
  author = {Jain, Anubhav and Ong, Shyue Ping and Hautier, Geoffroy and Chen, Wei and Richards, William Davidson and Dacek, Stephen and Cholia, Shreyas and Gunter, Dan and Skinner, David and Ceder, Gerbrand and Persson, Kristin A.},
  year = {2013},
  month = jul,
  journal = {APL Materials},
  volume = {1},
  number = {1},
  pages = {011002},
  doi = {10.1063/1.4812323},
  url = {https://aip.scitation.org/doi/10.1063/1.4812323},
  urldate = {2021-12-08}
}

@article{jain_highthroughput_2011,
  title = {A High-Throughput Infrastructure for Density Functional Theory Calculations},
  author = {Jain, Anubhav and Hautier, Geoffroy and Moore, Charles and Ong, Shyue and Fischer, Chris and Mueller, Tim and Persson, Kristin and Ceder, Gerbrand},
  year = {2011},
  month = jun,
  journal = {Comput. Mater. Sci.},
  volume = {50},
  pages = {2295--2310},
  doi = {10.1016/j.commatsci.2011.02.023}
}

@misc{jiao_crystal_2023,
  title = {Crystal {{Structure Prediction}} by {{Joint Equivariant Diffusion}}},
  author = {Jiao, Rui and Huang, Wenbing and Lin, Peijia and Han, Jiaqi and Chen, Pin and Lu, Yutong and Liu, Yang},
  year = {2023},
  month = jul,
  journal = {arXiv.org},
  url = {https://arxiv.org/abs/2309.04475v2},
  urldate = {2024-10-03},
  langid = {english}
}

@misc{jiao_space_2024,
  title = {Space {{Group Constrained Crystal Generation}}},
  author = {Jiao, Rui and Huang, Wenbing and Liu, Yu and Zhao, Deli and Liu, Yang},
  year = {2024},
  month = apr,
  number = {arXiv:2402.03992},
  eprint = {2402.03992},
  primaryclass = {cond-mat},
  publisher = {arXiv},
  doi = {10.48550/arXiv.2402.03992},
  url = {http://arxiv.org/abs/2402.03992},
  urldate = {2024-09-26},
  archiveprefix = {arXiv}
}

@misc{joshi_allatom_2025,
  title = {All-Atom {{Diffusion Transformers}}: {{Unified}} Generative Modelling of Molecules and Materials},
  shorttitle = {All-Atom {{Diffusion Transformers}}},
  author = {Joshi, Chaitanya K. and Fu, Xiang and Liao, Yi-Lun and Gharakhanyan, Vahe and Miller, Benjamin Kurt and Sriram, Anuroop and Ulissi, Zachary W.},
  year = {2025},
  month = mar,
  number = {arXiv:2503.03965},
  eprint = {2503.03965},
  primaryclass = {cs},
  publisher = {arXiv},
  doi = {10.48550/arXiv.2503.03965},
  url = {http://arxiv.org/abs/2503.03965},
  urldate = {2025-03-10},
  archiveprefix = {arXiv}
}

@article{kirklin_open_2015a,
  title = {The {{Open Quantum Materials Database}} ({{OQMD}}): Assessing the Accuracy of {{DFT}} Formation Energies},
  shorttitle = {The {{Open Quantum Materials Database}} ({{OQMD}})},
  author = {Kirklin, Scott and Saal, James E. and Meredig, Bryce and Thompson, Alex and Doak, Jeff W. and Aykol, Muratahan and R{\"u}hl, Stephan and Wolverton, Chris},
  year = {2015},
  month = dec,
  journal = {npj Comput Mater},
  volume = {1},
  number = {1},
  pages = {1--15},
  publisher = {Nature Publishing Group},
  issn = {2057-3960},
  doi = {10.1038/npjcompumats.2015.10},
  url = {https://www.nature.com/articles/npjcompumats201510},
  urldate = {2025-04-29},
  copyright = {2015 The Author(s)},
  langid = {english}
}

@inproceedings{levy_symmcd_2024,
  title = {{{SymmCD}}: {{Symmetry-Preserving Crystal Generation}} with {{Diffusion Models}}},
  shorttitle = {{{SymmCD}}},
  booktitle = {{{AI}} for {{Accelerated Materials Design}} - {{NeurIPS}} 2024},
  author = {Levy, Daniel and Panigrahi, Siba Smarak and Kaba, S{\'e}kou-Oumar and Zhu, Qiang and Galkin, Mikhail and Miret, Santiago and Ravanbakhsh, Siamak},
  year = {2024},
  month = nov,
  url = {https://openreview.net/forum?id=V7x2KZQn2v},
  urldate = {2024-12-10},
  langid = {english}
}

@misc{lipman_flow_2023,
  title = {Flow {{Matching}} for {{Generative Modeling}}},
  author = {Lipman, Yaron and Chen, Ricky T. Q. and {Ben-Hamu}, Heli and Nickel, Maximilian and Le, Matt},
  year = {2023},
  month = feb,
  number = {arXiv:2210.02747},
  eprint = {2210.02747},
  primaryclass = {cs, stat},
  publisher = {arXiv},
  doi = {10.48550/arXiv.2210.02747},
  url = {http://arxiv.org/abs/2210.02747},
  urldate = {2024-09-26},
  archiveprefix = {arXiv}
}

@article{merchant_scaling_2023,
  title = {Scaling Deep Learning for Materials Discovery},
  author = {Merchant, Amil and Batzner, Simon and Schoenholz, Samuel S. and Aykol, Muratahan and Cheon, Gowoon and Cubuk, Ekin Dogus},
  year = {2023},
  month = nov,
  journal = {Nature},
  volume = {624},
  pages = {80--85},
  publisher = {Nature Publishing Group},
  issn = {1476-4687},
  doi = {10.1038/s41586-023-06735-9},
  url = {https://www.nature.com/articles/s41586-023-06735-9},
  urldate = {2023-11-29},
  copyright = {2023 The Author(s)},
  langid = {english}
}

@misc{miller_flowmm_2024,
  title = {{{FlowMM}}: {{Generating Materials}} with {{Riemannian Flow Matching}}},
  shorttitle = {{{FlowMM}}},
  author = {Miller, Benjamin Kurt and Chen, Ricky T. Q. and Sriram, Anuroop and Wood, Brandon M.},
  year = {2024},
  month = jun,
  number = {arXiv:2406.04713},
  eprint = {2406.04713},
  primaryclass = {cond-mat, physics:physics, stat},
  publisher = {arXiv},
  doi = {10.48550/arXiv.2406.04713},
  url = {http://arxiv.org/abs/2406.04713},
  urldate = {2024-09-23},
  archiveprefix = {arXiv}
}

@misc{pickard_airss_2016,
  type = {Text},
  title = {{{AIRSS}}},
  author = {Pickard, Prof C. J.},
  year = {2016},
  month = oct,
  url = {https://www.mtg.msm.cam.ac.uk/Codes/AIRSS},
  urldate = {2025-05-07},
  langid = {english}
}

@article{pickard_initio_2011,
  title = {{\emph{Ab Initio}} Random Structure Searching},
  author = {Pickard, Chris J and Needs, R J},
  year = {2011},
  month = feb,
  journal = {J. Phys.: Condens. Matter},
  volume = {23},
  number = {5},
  pages = {053201},
  issn = {0953-8984, 1361-648X},
  doi = {10.1088/0953-8984/23/5/053201},
  url = {https://iopscience.iop.org/article/10.1088/0953-8984/23/5/053201},
  urldate = {2024-05-27},
  langid = {english}
}

@misc{song_scorebased_2021,
  title = {Score-{{Based Generative Modeling}} through {{Stochastic Differential Equations}}},
  author = {Song, Yang and {Sohl-Dickstein}, Jascha and Kingma, Diederik P. and Kumar, Abhishek and Ermon, Stefano and Poole, Ben},
  year = {2021},
  month = feb,
  number = {arXiv:2011.13456},
  eprint = {2011.13456},
  primaryclass = {cs},
  publisher = {arXiv},
  doi = {10.48550/arXiv.2011.13456},
  url = {http://arxiv.org/abs/2011.13456},
  urldate = {2025-01-07},
  archiveprefix = {arXiv}
}

@misc{togo_spglib_2024,
  title = {Spglib: A Software Library for Crystal Symmetry Search},
  shorttitle = {\${\textbackslash}texttt\{\vphantom\}{{Spglib}}\vphantom\{\}\$},
  author = {Togo, Atsushi and Shinohara, Kohei and Tanaka, Isao},
  year = {2024},
  month = mar,
  number = {arXiv:1808.01590},
  eprint = {1808.01590},
  primaryclass = {cond-mat},
  publisher = {arXiv},
  doi = {10.48550/arXiv.1808.01590},
  url = {http://arxiv.org/abs/1808.01590},
  urldate = {2024-07-05},
  archiveprefix = {arXiv}
}

@misc{wood_uma_2025,
  title = {{{UMA}}: {{A Family}} of {{Universal Models}} for {{Atoms}}},
  shorttitle = {{{UMA}}},
  author = {Wood, Brandon M. and Dzamba, Misko and Fu, Xiang and Gao, Meng and Shuaibi, Muhammed and {Barroso-Luque}, Luis and Abdelmaqsoud, Kareem and Gharakhanyan, Vahe and Kitchin, John R. and Levine, Daniel S. and Michel, Kyle and Sriram, Anuroop and Cohen, Taco and Das, Abhishek and Rizvi, Ammar and Sahoo, Sushree Jagriti and Ulissi, Zachary W. and Zitnick, C. Lawrence},
  year = {2025},
  month = jun,
  number = {arXiv:2506.23971},
  eprint = {2506.23971},
  primaryclass = {cs},
  publisher = {arXiv},
  doi = {10.48550/arXiv.2506.23971},
  url = {http://arxiv.org/abs/2506.23971},
  urldate = {2025-08-01},
  archiveprefix = {arXiv},
  langid = {english}
}

@misc{xie_crystal_2022,
  title = {Crystal {{Diffusion Variational Autoencoder}} for {{Periodic Material Generation}}},
  author = {Xie, Tian and Fu, Xiang and Ganea, Octavian-Eugen and Barzilay, Regina and Jaakkola, Tommi},
  year = {2022},
  month = mar,
  number = {arXiv:2110.06197},
  eprint = {2110.06197},
  primaryclass = {cond-mat, physics:physics},
  publisher = {arXiv},
  url = {http://arxiv.org/abs/2110.06197},
  urldate = {2022-10-31},
  archiveprefix = {arXiv}
}

@article{zeni_generative_2025,
  title = {A Generative Model for Inorganic Materials Design},
  author = {Zeni, Claudio and Pinsler, Robert and Z{\"u}gner, Daniel and Fowler, Andrew and Horton, Matthew and Fu, Xiang and Wang, Zilong and Shysheya, Aliaksandra and Crabb{\'e}, Jonathan and Ueda, Shoko and Sordillo, Roberto and Sun, Lixin and Smith, Jake and Nguyen, Bichlien and Schulz, Hannes and Lewis, Sarah and Huang, Chin-Wei and Lu, Ziheng and Zhou, Yichi and Yang, Han and Hao, Hongxia and Li, Jielan and Yang, Chunlei and Li, Wenjie and Tomioka, Ryota and Xie, Tian},
  year = {2025},
  month = jan,
  journal = {Nature},
  pages = {1--3},
  publisher = {Nature Publishing Group},
  issn = {1476-4687},
  doi = {10.1038/s41586-025-08628-5},
  url = {https://www.nature.com/articles/s41586-025-08628-5},
  urldate = {2025-01-16},
  copyright = {2025 The Author(s), under exclusive licence to Springer Nature Limited},
  langid = {english}
}

@misc{zeni_mattergen_2023,
  title = {{{MatterGen}}: A Generative Model for Inorganic Materials Design},
  shorttitle = {{{MatterGen}}},
  author = {Zeni, Claudio and Pinsler, Robert and Z{\"u}gner, Daniel and Fowler, Andrew and Horton, Matthew and Fu, Xiang and Shysheya, Sasha and Crabb{\'e}, Jonathan and Sun, Lixin and Smith, Jake and Tomioka, Ryota and Xie, Tian},
  year = {2023},
  month = dec,
  number = {arXiv:2312.03687},
  eprint = {2312.03687},
  primaryclass = {cond-mat},
  publisher = {arXiv},
  doi = {10.48550/arXiv.2312.03687},
  url = {http://arxiv.org/abs/2312.03687},
  urldate = {2023-12-08},
  archiveprefix = {arXiv}
}

\newpage
\appendix

\section{Equivariant Model}\label{app:CSPNet}

For the equivariant graph neural network utilized with OMatG-FC, we provide greater detail about its architecture here.
Node embeddings $\bm{h}^i_{(s)}$ are initialized as a function of the atom types $\bm{a}$ and updated through a message-passing scheme in which nearby node embeddings, the full lattice information $\bm{L}$, and sinusoidal distance embeddings are injected. This scheme is given by
\begin{align}
\begin{split}
    \bm{h}^i_{(0)} &= \phi_{\bm{h}_{(0)}} (\bm{a}_i) \\
    \bm{m}_{(s)}^{ij} &= \varphi_m \pt{ \bm{h}^i_{s-1}, \bm{h}^j_{s-1}, \bm{l}, \text{SinEmb}\pt{\bm{x}^j - \bm{x}^i}} \\
    \bm{m}_{(s)}^i &= \sum_{j=1}^N \bm{m}^{ij}_{(s)} \\
    \bm{h}_{(s)}^i &= \bm{h}_{(s-1)}^i + \varphi_h\pt{\bm{h}_{(s-1)}, \bm{m}_{(s)}^i} \\
    b_{V} &= \varphi_{V}\pt{\bm{h}^i_{(\mathrm{max \,} s)}} \\
    b_{\bm{A}} &= \varphi_{\bm{A}}\pt{\bm{h}^i_{(\mathrm{max \,} s)}}
    \label{eq:CSPNet}
\end{split}
\end{align}
wherein $\varphi_m$ is a parameterized neural network.
All necessary drift $b^\theta(t,x)$ and denoiser $z^\theta(t,x)$ terms are computed from linear layers applied to the hidden final layer node embeddings (of dimension 1 for the volume scale and dimension 101 for the atomic species). 
The network also predicts $\log p_{1|t}^\theta(\bm{a}_1|x_{t})$ for the atomic species.

\section{Hyperparameter Tuning}\label{app:hyperparams}

We perform hyperparameter tuning of the models based on the linear interpolant for both ODE and SDE sampling schemes, using the \texttt{Ray Tune} package \cite{liaw_tune_2018} with the Optuna library \cite{akiba_optuna_2019}. For training, we tune the relative loss weights $\lambda$, the batch size, the learning rate, the weight decay and, $\gamma(t)$ (for SDE). For inference, we tune the number of integration steps, the species noise $\eta$, the time-dependent noise $\varepsilon(t)$ (for SDE), and the magnitude of the velocity annealing parameter $s$ for the volume scale. We minimize with respect to the following evaluation metric:

\begin{align}
    \begin{split}
        \text{eval}_\text{FCP} = \text{avg} \Big[&1-\text{combined validity}, \\
        &\text{wdist}(\rho), \text{wdist}(N\text{ary}), \text{wdist}(\langle CN \rangle_S), \\
        &\text{wdist}(\text{elem}), \text{wdist}(\text{space group})\Big].
    \end{split}
\end{align}

\subsection{Hyperparameters}

We include the hyperparameters for the OMatG-FC models provided in this work below.

 \begin{table}[h]
\centering
\caption{ODE sampling scheme training hyperparameters.}
\label{tab:hyperparameters_fs}
\begin{tabular}{ll}
\toprule
\textbf{Hyperparameter} & \textbf{Value} \\
\midrule
\multicolumn{2}{l}{\textit{Optimization}} \\
Optimizer & Adam \\
Learning rate & $1.596 \times 10^{-4}$ \\
LR scheduler & None \\
Batch size & 128 \\
Max epochs & 1000 \\
Gradient clipping (value) & 0.5 \\
Precision & FP32 \\
Early stopping patience & 10 \\
\midrule
\multicolumn{2}{l}{\textit{Stochastic Interpolants}} \\
Integration time steps & 610 \\
Integration method & Euler \\
Equation type & ODE \\
Velocity annealing factor & 6.020 \\
DFM mask noise & 46.375 \\
Interpolant & Linear \\
\midrule
\multicolumn{2}{l}{\textit{Loss Weights}} \\
Species loss & 1.0 \\
Volume scale loss ($b$) & 0.4160 \\
\bottomrule
\end{tabular}
\end{table}
\begin{table}[h]
\centering
\caption{SDE sampling scheme training hyperparameters.}
\label{tab:hyperparameters}
\begin{tabular}{ll}
\toprule
\textbf{Hyperparameter} & \textbf{Value} \\
\midrule
\multicolumn{2}{l}{\textit{Optimization}} \\
Optimizer & AdamW \\
Learning rate & $8.313 \times 10^{-4}$ \\
Weight decay & $1.935 \times 10^{-5}$ \\
LR scheduler & CosineAnnealingLR \\
$T_{\max}$ & 1000 \\
$\eta_{\min}$ & $1 \times 10^{-7}$ \\
Batch size & 256 \\
Max epochs & 1000 \\
Gradient clipping (value) & 0.5 \\
Precision & FP32 \\
Early stopping patience & 10 \\
\midrule
\multicolumn{2}{l}{\textit{Stochastic Interpolants}} \\
Integration time steps & 620 \\
Integration step size $dt$ & $1.408 \times 10^{-3}$ \\
Integration method & Euler \\
Equation type & SDE \\
Velocity annealing factor & 1.805 \\
DFM mask noise & 25.710 \\
Interpolant & Linear \\
$c$ Vanishing Epsilon & 0.5303 \\
$\mu$ Vanishing Epsilon & 0.1749 \\
$\sigma$ (Vanishing Epsilon) & 0.4439 \\
$a$ (Latent Gamma (Sqrt)) & 0.02812 \\
\midrule
\multicolumn{2}{l}{\textit{Loss Weights}} \\
Species loss & 1.0 \\
Volume scale loss ($b$) & 0.9198 \\
Volume scale loss ($z$) & 0.2406 \\
\bottomrule
\end{tabular}
\end{table}

\clearpage

\section{Computing Resources}\label{app:compute}

All experiments were run on a single NVIDIA A100 GPUs. 
Training a single model on MP-20 for 1000 epochs takes $\sim$24 hours.
Hyperparameter tuning was conducted using 4 A100 GPUs and 16 CPUs for $\sim$96 hours.
Inference (sampling 10,000 structures) takes roughly 45 minutes with the same hardware.
Additionally, performing structural comparisons on the full $MP$-20 dataset took $\sim$2000 CPU hours.

\section{StructureMatcher}\label{app:SM}

Pymatgen's \texttt{StructureMatcher} is a commonly used tool to compute match rates between generated and reference structures. 
The algorithm determines whether two structures match using three tolerances: the lattice (\texttt{ltol}) and angle tolerance (\texttt{angle\_tol}), which control the allowed deviation in lattice lengths and angles, respectively; and the site tolerance, \texttt{stol}, which controls the allowed volume-normalized atomic-site displacements based on the average free length per atom, $(V/N)^\frac{1}{3}$, where $V$ is the unit cell volume and $N$ is the number of atoms. 
Matching typically proceeds by reducing structures to primitive cells (we however, set this to False), aligning lattice vectors within tolerances, and computing a normalized per-atom root-mean-square-distance (RMSD) between atomic positions.

\section{Data and code availability}

Code accompanying this paper will be made available at: \texttt{https://github.com/FERMat-ML/OMatG}. \\

The framework-split dataset discussed in Appendix~\ref{app:deduplication} will be made available at: \texttt{https://huggingface.co/colabfit}.

\clearpage

\section{Framework-split Dataset}\label{app:framework_splits}

\subsection{Framework Duplicate Analysis}\label{app:deduplication}

\begin{figure*}[ht]
   \centering
   \includegraphics[width=0.8\textwidth]{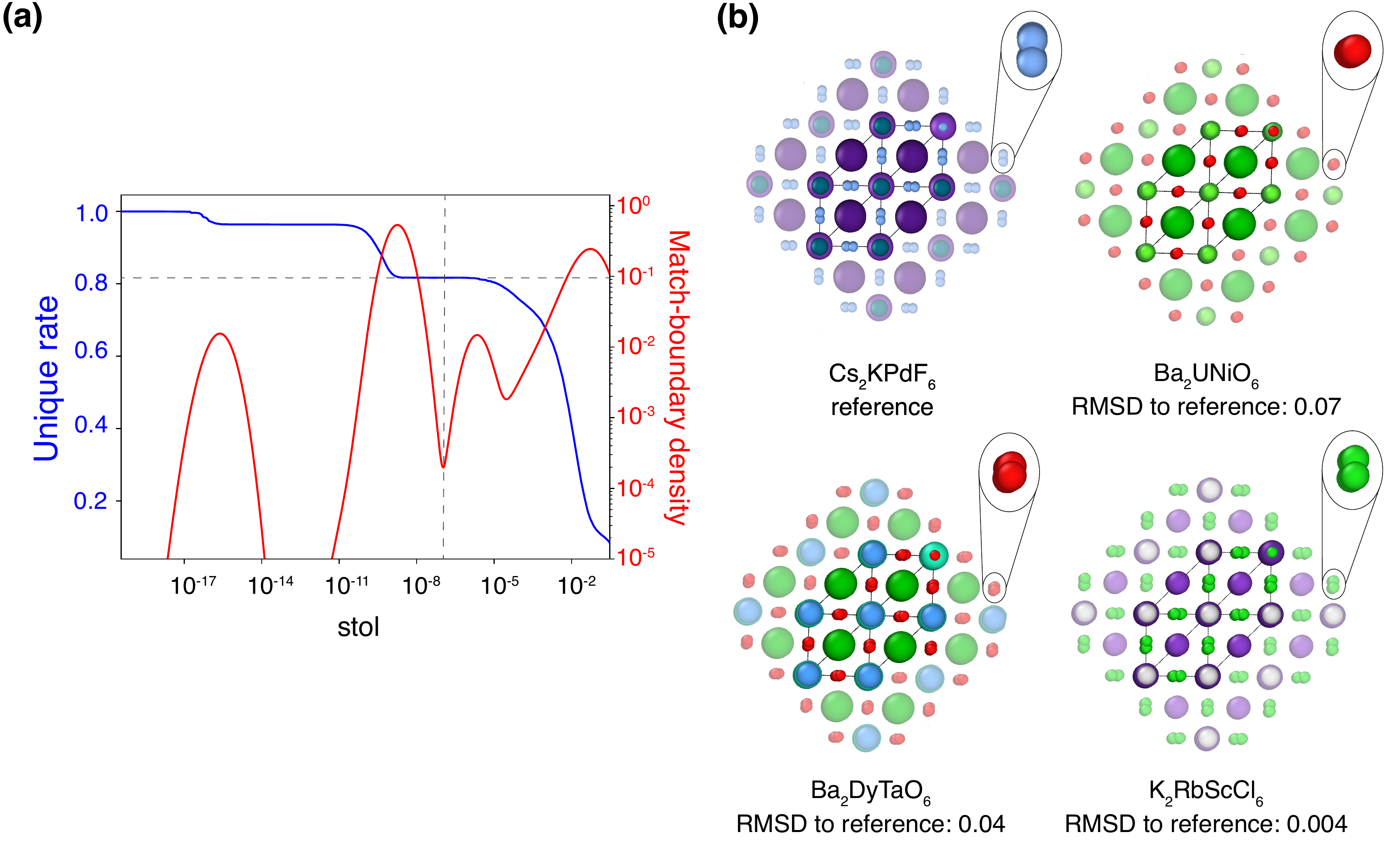}
   \caption{\textbf{(a)} Uniqueness rate and kernel density estimate (using a Gaussian kernel) of the match-boundary density as a function of \texttt{stol} of \texttt{StructureMatcher} plotted for the entire $MP$-20 dataset.
   \textbf{(b)} Examples of three structural frameworks with different compositions which `match' by framework to the reference Cs$_2$KPdF$_6$ structure, with the RMSDs and structural differences on one of the symmetry sites magnified for each.}
   \label{fig:framework_replicas}
\end{figure*}

We investigate structural framework duplicates within $MP$-20, demonstrating that frameworks are commonly shared across different materials---motivating our framework-constrained approach. 
Structure pair comparisons are made to find the \texttt{stol} tolerance at which they transition from matching to non-matching, which is equivalent to the RMSD between atomic positions at this transition. 

All pairs of structures are compared to one another using pymatgen's \texttt{StructureMatcher} with the default tolerances: the site tolerance \texttt{stol}$=0.3$, lattice length tolerance of \texttt{ltol}$=0.2$ and angle tolerance of \texttt{angle\_tol}$=5.0$. 
For every pair $(i,j)$ of the $N$ structures in $\mathrm{train}\cup\mathrm{val}\cup\mathrm{test}$, we compute the smallest site tolerance $s_{ij}\in[0, \texttt{stol}]$ at which pymatgen's \texttt{StructureMatcher} declares the two frameworks equivalent: This is the RMSD output by the matcher.  
Several non-standard arguments are added: the \texttt{primitive\_cell} argument is set to False to avoid reduction to primitive cells for matching; \texttt{scale} is set to True, allowing for identical structural frameworks with different volume scales to match; and the \texttt{FrameworkComparator} ignores the identity of the chemical species. 
Our setup thus compares the structural frameworks only.
Pairs that fail to match at \texttt{stol} are assigned $s_{ij}=\texttt{stol}$ as a penalty; smaller $s_{ij}$ indicates a closer match. The comparisons are evaluated and stored in an upper-triangular matrix which is later also utilized for creating framework splits according to Appendix~\ref{app:fs-code}.

Figure~\ref{fig:framework_replicas} reveals large peaks in match-boundary density at \texttt{stol}~$\approx \SI{1e-9}{}$ and ~$\approx \SI{1e-16}{}$, identifying exact framework duplicates. 
These structures differ only by what we posit to be from dataset preparation---numerical noise (possibly from structure symmetrization) or stopping criteria for DFT-based structure relaxation. These duplicates account for ${\sim}18\%$ of $MP$-20 materials. 
The match boundary density also reveals multiple valleys (one at \SI{1e-7}{} and another at \SI{1e-4}{}) which would both suffice as duplicate cutoff values of \texttt{stol}, with only a modest difference in the fraction of unique structures between them.
At \texttt{stol} tolerances above \SI{1e-4}, structures are strikingly similar, but not duplicates.
Visual inspection (Fig.~\ref{fig:framework_replicas}b) illustrates how \ce{Ba2DyTaO6} and \ce{K2RbScCl6} share similar geometric motifs as the reference \ce{Cs2KPdF6} with completely different chemistries and slightly varying structure.
It is worth noting, however, that at the default tolerance for \texttt{StructureMatcher} of \texttt{stol}$=0.3$, only ${\sim}10\%$ structural frameworks would be considered unique.

\subsection{Performing Framework Splits}\label{app:fs-code}

The procedure for creating a framework-split dataset consists of two stages: first, construction of a pairwise framework-similarity matrix as described in Appendix~\ref{app:deduplication}, followed by a similarity-respecting assignment of clusters of similar structures to splits \textit{via} hierarchical clustering and bin-packing.

We define the similarity threshold to be $\texttt{stol}=0.3$. Connected components are found \textit{via} a Union–Find structure, with edges stored in a dictionary. 
Each component is subsequently refined using complete-linkage agglomerative clustering, with dendrograms cut at \texttt{stol}$=0.3$, ensuring that all within-cluster pairwise distances remain below \texttt{stol}$=0.3$. 
Clusters are assigned to splits using First-Fit Decreasing bin-packing with target ratios corresponding to the 60--20--20 splits: clusters are sorted by size and greedily allocated to the most underfilled split, with seeded tie-breaking for reproducibility.
Statistics about the framework-split dataset can be found in Appendix~\ref{app:fs-stats}.

\section{Comparisons to Prototype-Constrained Methods}\label{app:MGB}

We benchmark against two prototype-constrained methods---random enumeration and ionic substitution \cite{hautier_data_2011a}---utilizing the \texttt{matgen-baselines} codebase \cite{szymanski_2025_matgen_baselines}; results are shown in Tab.~\ref{tab:dng_eval}.
These methods modify only the atomic species on a given structure’s atomic “decoration” and leave the fractional coordinates and unit cell unchanged. 
We modify the existing code such that the output can be directly compared to the generative modeling task---while the existing code pulls candidate materials from the full Materials Project database and determines all valid combinations for each material, here we only generate one structure per $MP$-20 test set structure. 
For random enumeration, the first valid candidate meeting oxidation state criteria is stored. 
For ionic substitution, because valid candidates have differing probabilities based on how commonly the substitution occurs in known materials, we benchmark with only highest-probability candidate. 
Not all $MP$-20 test set structures are able to produce valid candidates, but for fairness of comparison we calculate all metrics as percentages out of 2500 input structures.

We summarize by highlighting three advantages of OMatG-FC that cannot be achieved using either of these methods: (1) the ability to take as input any periodic point pattern without any chemical decoration on it previously, (2) the ability to discover new prototypes based on the same structural framework (see Fig.~\ref{fig:prototype}), and (3) we are not limited to structures which have meaningful oxidation states, as is required for the ionic substitution method.

\clearpage

\section{Additional Performance Quantification}

\subsection{Volume-scaling ablations}

We test our model's performance when learning volume scaling is ablated, to isolate the value of our lattice factorization. 
We do so by training a new model that fixes the volume to that of the framework's original structure, using the $MP$-20 dataset to be directly comparable to Tab.~\ref{tab:dng_eval}.
Tab. ~\ref{tab:metrics_no_volume} shows the results of this ablation.
Compared to our model with volume learning, fixing the volume produces comparable (M)S.U.N. rates, with higher stability rates but lower novelty rates. 
We note, however, that framework-conditioning is a hard constraint only for unrelaxed structures.
For relaxation, all structures use the same UMA implementation with ASE's \texttt{FrechetCellFilter}, allowing atomic positions and all six independent cell-strain components---including anisotropic strain---to change.
Relaxation thus softens the framework constraint, though local optimization cannot guarantee reconstruction into a different framework basin. 


\begin{table*}[h]
    \caption{Evaluation of the OMatG-FC models with volume-scaling ablation, both before and after pre-relaxation. Benchmarks are performed on \num{2500} generated structures using LeMat-GenBench \citep{betala_lematgenbench_2026}.}
    \label{tab:metrics_no_volume}
    \vspace{3pt}
    \centering
    \resizebox{0.9\textwidth}{!}{
    \begin{tabular}{lcccccc}
        \toprule
        \multirow{2}{*}{Method}
        & Validity & Novelty & Stable & Metastable & S.U.N. & M.S.U.N. \\
        & Rate (\%, $\uparrow$)
        & Rate (\%, $\uparrow$)
        & Rate (\%, $\uparrow$)
        & Rate (\%, $\uparrow$)
        & Rate (\%, $\uparrow$)
        & Rate (\%, $\uparrow$) \\
        \midrule
        OMatG-FC-LinearODE, unrelaxed & 96.2 & 30.2 & 4.8 & 58.1 & 0.2 & 6.7 \\
        OMatG-FC-LinearSDE, unrelaxed & 96.4 & 32.2 & 4.2 & 55.4 & 0.1 & 6.2 \\
        OMatG-FC-LinearODE, relaxed & 96.8 & 29.1 & 19.1 & 58.2 & 1.3 & 12.9 \\
        OMatG-FC-LinearSDE, relaxed & 97.1 & 31.1 & 17.2 & 57.7 & 1.1 & 12.5 \\
        \bottomrule 
    \end{tabular}
    }
\end{table*}

\subsection{Framework-Split Dataset}\label{app:fs-bench}
Appendix~\ref{app:fs-code} describes how structural framework comparisons with pymatgen's \texttt{StructureMatcher} are used to create a framework-split version of the $MP$-20 dataset, where matching frameworks are grouped into the same training, validation, or test split (using a 60--20--20 \% split). 
Table~\ref{tab:metrics_split} collects results for OMatG-FC models trained on this re-split version of the dataset. 
We used the same hyperparameters as for the models trained on the original $MP$-20 dataset, and observe a modest decrease in performance, particularly for stability and validity (see Tab.~\ref{tab:dng_eval})---which is expected given the more challenging learning task.

\begin{table*}[h]
    \caption{Benchmarks for \textit{de novo} generation as provided by the LeMat-GenBench software \citep{betala_lematgenbench_2026} on \num{2500} generated structures as in Tab.~\ref{tab:dng_eval}. The OMatG-FC models in this table are trained on the framework-split $MP$-20 dataset instead of the original $MP$-20 dataset.}
 \label{tab:metrics_split}
 \vspace{3pt}
  \centering
  \resizebox{0.9\textwidth}{!}{
  \begin{tabular}
  {lccccc}
    \toprule
    \multirow{2}{*}{Method} 
    & Validity & Stable & Metastable & S.U.N. & M.S.U.N. \\
    & Rate (\%, $\uparrow$)
    & Rate (\%, $\uparrow$)
    & Rate (\%, $\uparrow$) 
    & Rate (\%, $\uparrow$) 
    & Rate (\%, $\uparrow$) \\
 \midrule
 OMatG-FC-LinearODE, unrelaxed & 88.9 & 1.6 & 13.6 & 0.1 & 4.9 \\
 OMatG-FC-LinearODE, relaxed & 91.3 & 3.8 & 34.9 & 0.7 & 19.1 \\
 OMatG-FC-LinearSDE, unrelaxed & 89.8 & 1.6 & 17.0 & 0.1 & 6.5 \\
 OMatG-FC-LinearSDE, relaxed & 91.6 & 3.8 & 36.7 & 0.8 & 19.8 \\
  \bottomrule
\end{tabular}
}
\end{table*}

\clearpage

\section{Dataset Framework Statistics}

\subsection{Statistics on the Framework-Split Dataset}\label{app:fs-stats}

We further provide statistics on the framework-split dataset resulting from the procedure for creating it outlined in Appendix~\ref{app:fs-code}.
We find that $MP$-20 contains 3,183 unique frameworks among 45,229 materials using the criteria $\texttt{stol}=0.3$, $\texttt{ltol}=0.2$, and $\texttt{angle\_tol}=5.0$ (the default settings for pymatgen's \texttt{StructureMatcher}), and using a comparator that ignores species. 
An average of 14.21 materials have the same framework, though this appears heavily skewed---the most common framework is shared by 4,815 materials, whereas the median framework is represented by only one material. 
Out of the 3,183 unique frameworks, 1,698 were found to correspond to only one material in the $MP$-20 dataset. The standard $MP$-20 test set contains 1,259 distinct frameworks out of 9,046 materials, of which 388 (~31\%) are absent from the training set. 
This means ~69\% of frameworks in the standard $MP$-20 test set have at least one representative material in the training set.

For our new framework-split dataset, we provide the number of unique frameworks and corresponding materials present in each set in Tab.\ref{tab:framework_split}.
Framework splits were determined using a greedy largest-first procedure, in which frameworks are sorted from largest to smallest and the next framework is assigned to the split with the largest remaining material deficit---which early on is the training set, meaning it receives the frameworks with the most duplicates.

\begin{table}[h]
    \caption{Number of unique frameworks and materials in each split of the framework-split $MP$-20 dataset.}
    \label{tab:framework_split}
    \vspace{3pt}
    \centering
    \begin{tabular}{lcc}
        \toprule
        Split & Frameworks & Materials \\
        \midrule
        Train & 1,067 & 27,137 \\
        Validation & 1,058 & 9,046 \\
        Test & 1,058 & 9,046 \\
        \midrule
        \textbf{Total} & \textbf{3,183} & \textbf{45,229} \\
        \bottomrule
    \end{tabular}
\end{table}

\subsection{Diversity of frameworks in \textit{de novo} generated structures}

To investigate the structural and chemical diversity of structures generated by our model, we group the \num{2500} structures generated by OMatG-FC-LinearODE into distinct frameworks (following the convention described in Appendix~\ref{app:framework_splits}) and count the number of unique compositions satisfying the M.S.U.N. or S.U.N. criteria within each framework category.
As shown in Tab.~\ref{tab:framework_diversity}, the \num{2500} generated structures span 539 distinct frameworks, of which 101 contain at least one (M.)S.U.N. structure.
While most of these frameworks are associated with only one (M.)S.U.N. composition, some support multiple distinct compositions, with up to 13 observed within a single framework.

\begin{table}[h]
    \caption{Distribution of distinct (M.)S.U.N. compositions across the structural frameworks identified among \num{2500} structures generated by OMatG-FC-LinearODE.}
    \label{tab:framework_diversity}
    \vspace{3pt}
    \centering
    \begin{tabular}{cc}
        \toprule
        \begin{tabular}{c}
            Distinct (M.)S.U.N. \\
            compositions per framework
        \end{tabular}
        &
        \begin{tabular}{c}
            Number of \\
            frameworks
        \end{tabular} \\
        \midrule
        0  & 438 \\
        1  & 73 \\
        2  & 15 \\
        3  & 3 \\
        4  & 4 \\
        5  & 1 \\
        6  & 2 \\
        7  & 1 \\
        10 & 1 \\
        13 & 1 \\
        \midrule
        \textbf{Total} & \textbf{539} \\
        \bottomrule
    \end{tabular}
\end{table}




\clearpage

\end{document}